# DFPENet-geology: A Deep Learning Framework for High Precision Recognition and Segmentation of Co-seismic Landslides

**Qingsong Xu[1,3], Chaojun Ouyang[1,2,3*], Tianhai Jiang[1], Xuanmei Fan[4], Duoxiang Cheng[5]**

*[1)] Key laboratory of Mountain Hazards and Surface Process, Institute of Mountain Hazards and Environment, Chinese Academy of Sciences, Chengdu, 610041, China*

*[2)] CAS Center for Excellence in Tibetan Plateau Earth Sciences, Chinese Academy of Sciences (CAS), Beijing 100101, China.*

*[3)] University of Chinese Academy of Sciences, Beijing 100049, China*

*[4)] State Key Laboratory of Geohazard Prevention and Geoenvironment Protection, Chengdu University of Technology, Chengdu 610059, China*

*[5)] Sichuan Engineering Research Center for Emergency Mapping & Disaster Reduction/Sichuan Geomatics Center, Chengdu 610041, China*

## Abstract:

Automatic recognition and segmentation methods now become the essential requirement in identifying co-seismic landslides, which are fundamental for disaster assessment and mitigation in large-scale earthquakes. This approach used to be carried out through pixel-based or object-oriented methods. However, due to the massive amount of remote sensing data, variations in different earthquake scenarios, and the efficiency requirement for post-earthquake rescue, these methods are difficult to develop into an accurate, rapid, comprehensive, and general (cross-scene) solution for co-seismic landslide recognition. This paper develops a robust model, Dense Feature Pyramid with Encoder-decoder Network (DFPENet), to understand and fuse the multi-scale features of objects in remote sensing images. The proposed method achieves a competitive segmentation accuracy on the public ISPRS 2D Semantic. Furthermore, a comprehensive and widely-used scheme is proposed for co-seismic landslide recognition, which integrates image features extracted from the DFPENet model, geologic features, temporal resolution, landslide spatial analysis, and transfer learning, while only RGB images are used. To corroborate its feasibility and applicability, the proposed scheme is applied to two earthquake-triggered landslides in Jiuzhaigou (China) and Hokkaido (Japan), using available pre- and post-earthquake remote sensing images. The experiments show that the proposed scheme presents a new state-of-the-art performance in regional landslide identification, and performs well in different seismic landslide recognition tasks, though landslide boundary error is not considered. The proposed scheme demonstrates a competitive performance for high-precision, high-efficiency and cross-scene recognition of earthquake disasters, which may serve as a starting point for the application of deep learning methods in co-seismic landslide recognition.

## 1. Introduction

Large-scale earthquakes can trigger a series of disasters, especially the co-seismic landslides (Huang et al., 2013; Keefer, 1984; Kjekstad and Highland, 2009), bringing tremendous damages to urban and rural areas around the world. To recognize co-seismic landslides and assess its risks are of great importance to emergency response and reconstruction and the development of a safe and resilient society, which brings considerable research attention to the field in recent years. Recent advancements in remote sensing and its applications in geological disaster make it possible to use remote sensing images for emergency response and rapid assessment of disasters. However, due to the massive amount of remote sensing data, the variations in different earthquake scenarios, and the time requirement for post-earthquake rescue, it is still difficult to develop an accurate, rapid, comprehensive, and general (cross-scene) method for seismic landslide recognition.

The methods for landslide recognition mainly include artificial interpretation and computer interpretation, with the latter serves as the main method for co-seismic landslide maps. For automatic extraction of remote sensing images, processing at the pixel level was adopted in the early stage, for example, Borghuis et al. (2007) used supervised and unsupervised classification to identify and map out landslide areas; Danneels et al. (2007) utilized a supervised pixel classification algorithm to detect landslides; Nichol and Wong (2005) showed how change detection technique was successfully applied to differentiate landslides. However, methods at the pixel level contain many issues, including the single expression of features, the lack of knowledge fusion, the isolation between remote sensing and GIS, among others. The above methods rely only on the spectral images. Then, multivariate models which combine spectral images with other data were proposed by researchers. For instance, topographic data of landslides (Iwahashi and Pike, 2007), panchromatic images (Van Westen et al., 2008), and other forms of data, are taken into consideration for landslide detection. Afterwards, a knowledge driven method, Object-Oriented Analysis (OOA), rose in the field of landslide recognition through remote sensing images. A considerable amount of literature about OOA has been published, for instance, spectral, spatial and morphometric properties of landslides were used by Martha et al. (2010); texture feature in panchromatic images was used by Martha et al. (2012); geometrical and spectral characteristics of landslides were used by Ma et al. (2016), etc. This method is easy to be implemented, with low computational complexity, high iteration efficiency, and solid noise resistance. However, it often fails to recognize small objects in face of regional landslides, while the size of the search area during iterative processes will significantly affect on the accuracy and efficiency of the algorithm.

More recently, deep-learning methods have shown a dominant performance in landslide recognition, especially Deep Convolutional Neural Networks (DCNNs), which has strong feature expressions as well as learning and distinguishing abilities (Krizhevsky et al., 2012). This method requires a large number of samples to improve training results, and it can benefit from the large amount of remote sensing image data. Several attempts have been made to automatically recognize landslides by DCNNs (Ghorbanzadeh et al., 2019; Liu and Wu, 2016; H. Yu et al., 2017). In addition, many change detection methods based on DCNNs were used to find the changed areas (Z. Chen et al., 2018; Lei et al., 2019; Rudner et al., 2019). These methods have achieved fairly good results in landslide recognition with a huge number of training samples. However, these methods were only applied in specific research areas, for instance, Ding et al. (2016) evaluated its DCNN method by regional landslide events in Shenzhen, China; the work of Ghorbanzadeh et al. (2019) covered the landslides in the higher Himalayas. Once cross-scenes and multi-sensors are considered, the accuracy of the model will be reduced significantly. Thus, it is necessary to build a learnable and life-long model to perform recognition tasks in different seismic landslides. In addition, regional co-seismic landslide recognition based on DCNNs has been a largely under-explored domain, and the accuracy of landslide recognition needs to be further improved. The average accuracy of regional seismic landslide recognition based on the DCNNs methods mentioned above is around 70%. More importantly, these methods cannot distinguish small landslides well enough due to lack of processing abilities in more useful low-level features in DCNNs. Moreover, DCNN models can only extract remote sensing image features, while landslide geological conditions are barely considered. Thus, it is necessary to integrate geologic features into DCNNs, in order to further improve the accuracy of co-seismic landslide recognitions.

To tackle the above problems, a comprehensive and widely-used scheme for recognition of co-seismic landslides is proposed in this paper. This approach aims to achieve high efficiency, high precision, cross-scene, and easy operability of co-seismic landslides recognition. First, we propose a novel end-to-end network for semantic segmentation in remote sensing images, based on the encoder-decoder with dense feature pyramid network, named as the Dense Feature Pyramid with Encoder-decoder Network (DFPENet). The network is to understand and fuse the multi-scale features of objects in remote sensing images. A new feature filter module in the network is designed to control information propagation, using soft-attention gate mechanism and gated convolution network, while redundancies can also be effectively minimized in the network. A dense feature pyramid module is then utilized to gather more contextual information. The ISPRS Vaihingen (ISPRS, 2016) is adopted to evaluate the proposed network due to the lack of standard dataset for seismic landslide recognition and segmentation. The local experimental results demonstrate that DFPENet outperforms other state-of-the-art DCNN-based models and advanced methods on the benchmark.

Next, a segmentation scheme for seismic landslides based on DFPENet is designed. In order to integrate geologic features into DFPENet, we build a geologic feature fusion module by morphological features of landslides, which is called the DFPENet-geology. In addition, the temporal resolution of the seismic remote sensing images is added into the DFPENet-geology model, to achieve high-precision recognition results of the seismic landslides. The segmentation results are then vectorized and set to be editable to further analyze the spatial relationships among landslides in the region. Lastly, we transfer the model to perform new recognition tasks which only contain limited samples.

To test its accuracy and efficiency, the proposed scheme is applied to the earthquake-triggered Jiuzhaigou landslides. The experiment results show that the proposed model sets a new state-of-the-art performance in regional landslide recognition, without considering the landslide boundary error. Furthermore, we take the earthquake-triggered Hokkaido landslides in Japan as an example to test the applicability of transfer learning method. The result shows that the scheme performs well in different seismic landslide tasks. The proposed scheme has managed to develop a robust model for DCNNs research in remote sensing, and may contribute to the rapid and accurate recognition of future seismic landslides.

## 2. Related work

### 2.1. Semantic segmentation in remote sensing

Semantic segmentation, with the goal of assigning semantic labels to every pixel in an image (Mottaghi et al., 2014), is one of the basic tasks in computer vision. It plays a vital role in many important applications of remote sensing (Bruzzone and Demir, 2014; Zhang et al., 2016). With the development of deep learning method on remote sensing images, end-to-end segmentation in remote sensing becomes the mainstream. Sun and Wang (2018) establish a semantic segmentation scheme, which combines with DSM, for remote sense images. The scheme is based on Fully Convolutional Networks (FCN) (Long et al., 2015). Wang et al. (2017a) propose a gated network, based on the information entropy of the feature maps. The method can effectively integrate with local details and contextual information. Cheng et al. (2017) utilizes the cascaded convolutional neural network to automatically detect roads, which is the first attempt to use one cascaded network to bridge two subtasks together in the remote sensing applications. Liu et al. (2018a) proposes a self-cascaded network to successively aggregate contexts from large scale to small ones. Cheng et al. (2019) designs a Context Fuse Module to exploit contextual information. For all the above methods, it is critical in exploring how to build an effective mechanism to capture more contextual features of the objects in remote sense images.

### 2.2. Context-reinforced semantic segmentation

Context is a basic processing module in semantic segmentation. It is considered as one of the important factors in using multi-level feature fusion to reinforce contextual features in the semantic segmentation model. FCN (Long et al., 2015) and U-Net (Ronneberger et al., 2015) fuse information from lower layers through skip-connections. Deeplab series (Chen et al., 2017; Chen et al., 2018b, 2018a) develop Atrous Spatial Pyramid Pooling (ASPP) to capture multi-scale feature by dilated convolutional layers. Feature Pyramid Networks (FPN) (Lin et al., 2016) is proposed to identify objects of different sizes, which combines bottom-down with top-down methods to obtain strong contextual information, improving the performance of target detection and semantic segmentation on multiple datasets. Dense top-down networks (Bilinski and Prisacariu, 2018a) is subsequently designed to fuse more multi-level features. In addition, gating mechanism and attention mechanism are also used in context-reinforced modules. Gates are commonly utilized in Long Short-Term

Memory (LSTM) to control information propagation, while a sigmoid layer is treated as a gate unit in convolution layers (Li et al., 2019; Xu et al., 2018; Zeng et al., 2018). Moreover, Nilsson and Nivre (2009) explore an attention mechanism by interpreting gradient of output class scores with respect to the input image. Attention mechanisms are divided into hard- and soft-attention mechanisms. Li et al. (2018) and Oktay et al. (2018) use a soft-attention mechanism to improve the recognition accuracy of small objects by summarizing all locations.

## 3. The proposed model

The Dense Feature Pyramid with Encoder-decoder Network (DFPENet) proposed by this paper is designed to meet the challenges for semantic segmentation in remote sensing of seismic landslides. Then, the segmentation scheme, combining DFPENet with geologic features, is presented in this paper for seismic landslide recognition. At the end of this section, a unified seismic landslide recognition and segmentation method is designed using transfer learning approaches.

### 3.1. Challenges of remote sensing segmentation

It is a challenging task for segmentation of remote sensing images, which is different from natural images. The major challenge is that the fine structure features of objects in remote sensing images with high intra-class variance and low inter-class variance are difficult to obtain, which brings difficulties to high-precision recognition. In addition, many targets (such as landslides) show various orientations, structures and boundary shapes. Therefore, it is necessary to better understand and integrate multi-scale features, while eliminate information redundancy, to improve the precision of remote sensing recognition.

### 3.2. Dense Feature Pyramid with Encoder-decoder Network (DFPENet)

Based on the main question of understanding and integrating multi-scale features in remote sensing target segmentation, this paper proposes a new approach of semantic segmentation, named the Dense Feature Pyramid with Encoder-decoder Network (DFPENet). The structure of DFPENet is summarized in Fig. 1, which consists of two parts: encoder and decoder. Encoder is to generate a feature pyramid (Lin et al., 2016) with different levels by the backbone network-ResNet101, with dilated convolution (Yu and Koltun, 2015) and Atrous Spatial Pyramid Pooling (ASPP) (Chen et al., 2018a, 2018b) maximizing the difference among different feature levels. Consequently, Low-level features pay more attention to local details, while high-level features focus more on the overall situation. In the decoder part, the objective is to obtain the useful information effectively, while eliminating irrelevant information by Attention Gate Mechanism (AGM) and Gated Convolution Networks (GCN); in addition, we further design a dense top-down feature pyramid module, to gather more contextual information from outputs of AGM, GCN and encoder. Detailed descriptions of DFPENet, including its optimization and its Vaihingen dataset evaluation, are presented in the following subsections. It is worth noticing that standard remote sensing dataset, the Vaihingen, is used to verify the effectiveness of the proposed network, due to the lack of standard dataset for seismic landslide recognition and segmentation.

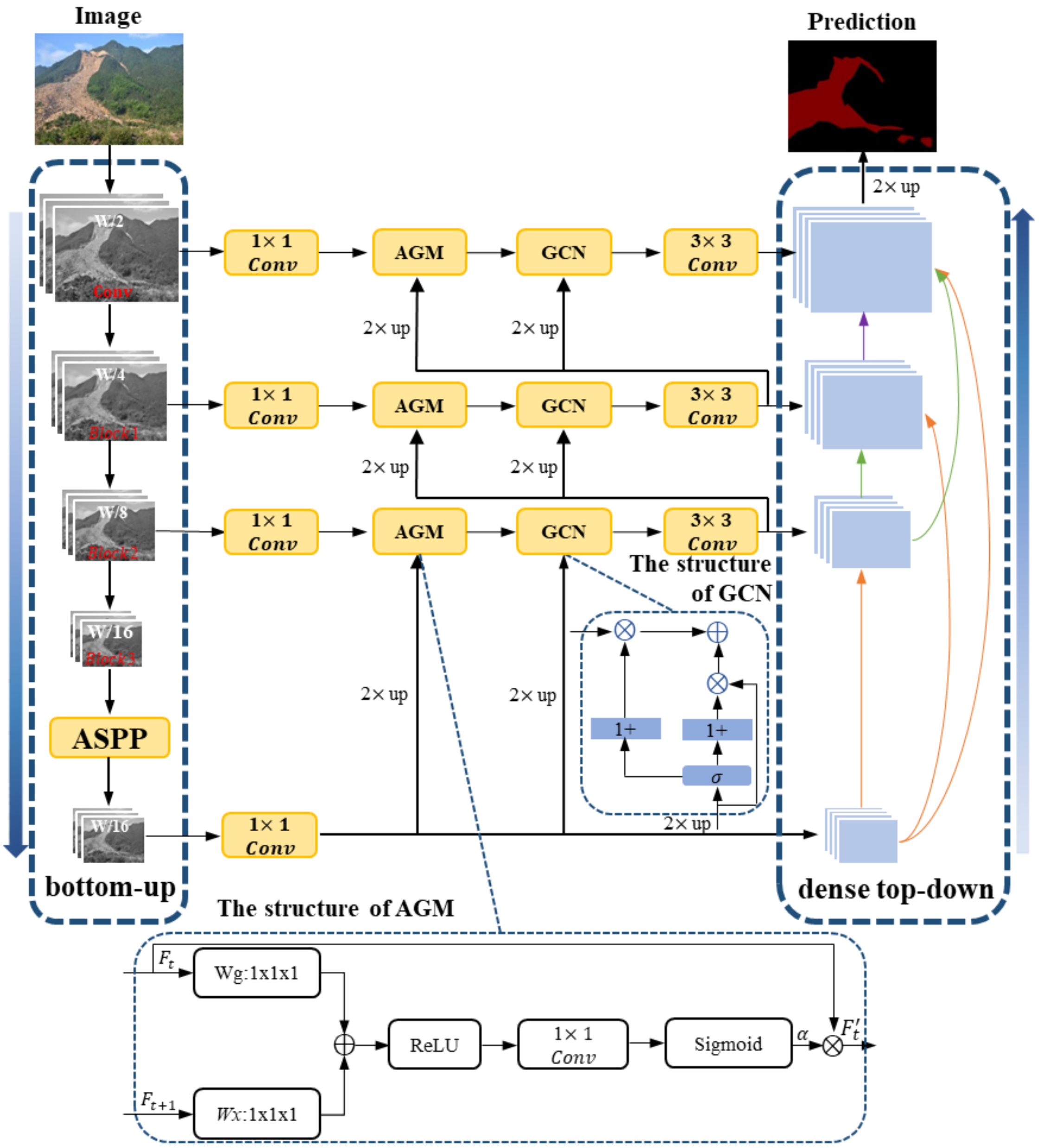

**Fig. 1.** The overview of our proposed Dense Feature Pyramid with Encoder-decoder Network. In the encoder part, ResNet-101 is used to extract features, while Dense top-down Feature Pyramid Module (DFPM) is designed to associate more contextual information in the decoder part. Attention Gate Mechanism (AGM) and Gated Convolution Networks (GCN) are then proposed for filter features. The structures of AGM and GCN are shown in the dashed boxes.

### 3.2.1. Bottom-up feature pyramid module

The role of bottom-up feature pyramid module is to generate multi-scale features from remote sensing images. In ResNet-101 (He et al., 2016), subsampling layers are necessary to enlarge the size of the receptive field, but they will decrease the spatial resolution of feature maps. Inspired by Dilated Resnet (F. Yu et al., 2017a) and Deeplabv3+ (Chen et al., 2018), we have adopted dilated convolutions (Yu et al., 2017) in the ResNet101 (He et al., 2016) to satisfy both the larger receptive field and the higher spatial resolution, shown in Fig. 2. Compared with the ResNet-101 in Deeplabv3+ (Chen et al., 2018b), we apply the dilate rates=[1,2,5] to the last three blocks in Fig. 2, which is better for feature extraction validated by experiments. In addition, the same operation as the Deeplab (Chen et al., 2018a, 2018b) utilized in this paper is that, Atrous Spatial Pyramid Pooling (ASPP) module is applied before the encoder output, in order to explore multi-scale features. Moreover, output stride=16 have been adopted to obtain as many contextual features as possible. As a result, the feature map of the encoder output contains 256 channels and rich context information.

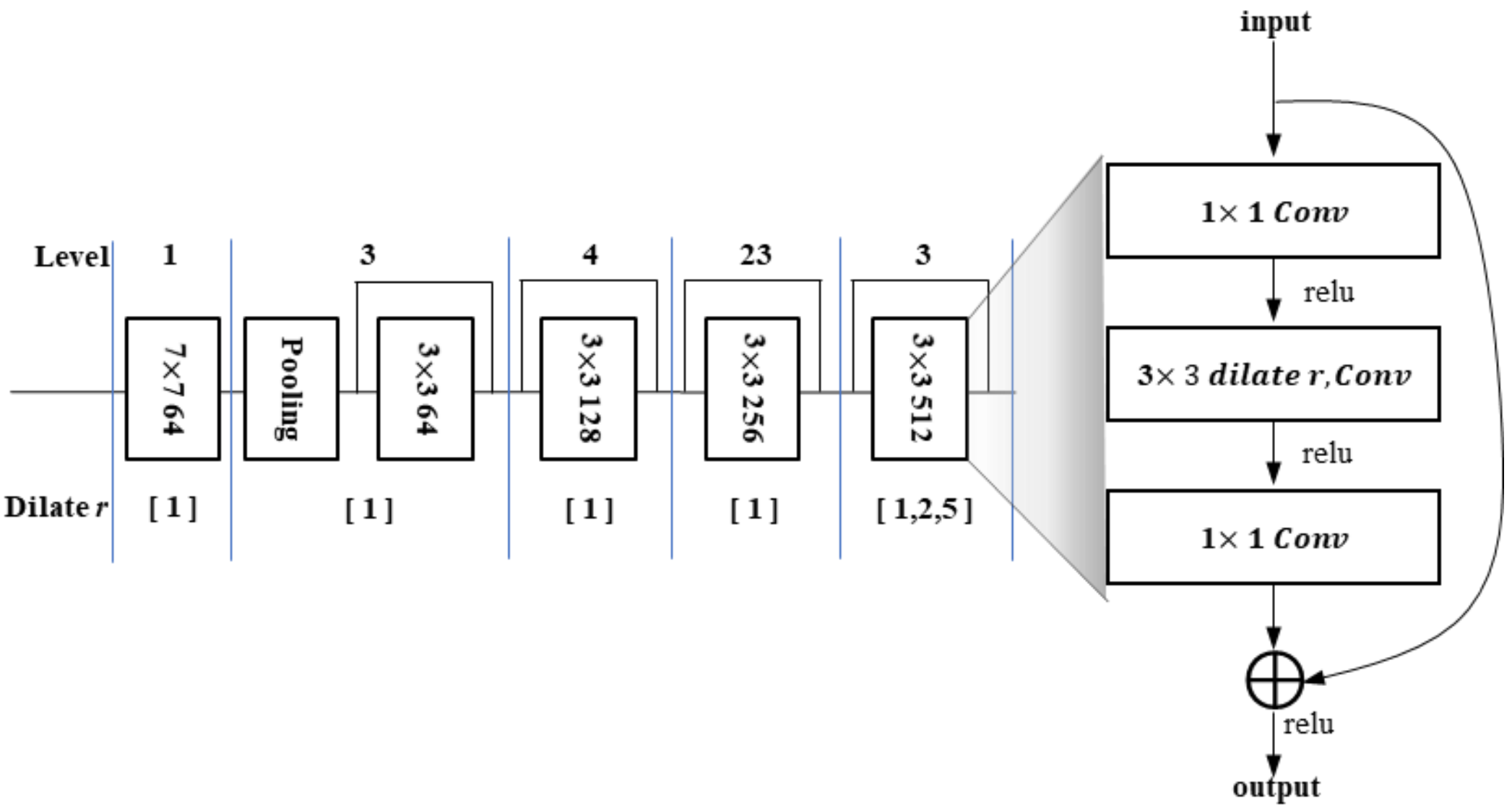

**Fig.2** The structure of ResNet101 with dilated convolutions. The dilate rates=1,2,5 are applied in the last three blocks.

### 3.2.2. Feature filter module

The core function of feature filter is to bring useful information about prediction to the right place, avoiding information redundancy. Attention Gate Mechanism (ACM) and Gated Convolution Networks (GCN) are the basic modules of feature filter. Specifically, AGM is one of the soft-attention techniques(Xu et al., 2015) to improve the recognition accuracy of small objects with vague boundary shape in remote sensing. For instance, small landslides (Volume $< 10 \times 10^4 m^3$) might be only about 20 to 30 pixels in thousands of pixels in seismic remote sensing images. The structure of the proposed ACM is shown in Fig. 1. GCN module is designed to control information flow with gates, which is based on the weighted sum of features, shown at the bottom of Fig. 1. The feature filter strategies can be formulated as,

$$F_t = (1+G_{t+1})\bullet F_{t+1} + (1+G_{t+1})\bullet A_t \bullet F_t \tag{1}$$

$$A_t = \sigma_2(\varphi^T(\sigma_1(W_g^T * F_t + W_x^T * F_{t+1}))) \tag{2}$$

$$G_{t+1} = \sigma_2(W_{t+1} * F_{t+1}) \tag{3}$$

Where $\sigma_1 = \max(0, x)$ and $\sigma_2 = \frac{1}{1+e^{-x}}$ correspond to ReLU and Sigmoid activation functions. The symbol $\bullet$ denotes element-wise multiplication along with the dimensions of feature channels. $*$ denotes channel-wise 1×1×1 convolutional layer parameterized with $W_x \in \mathbb{R}^{C_{t+1}\times C_{\text{int}}}$, $W_g \in \mathbb{R}^{C_t\times C_{\text{int}}}$, $W_g \in \mathbb{R}^{C_{t+1}\times 1}$. In addition, $\varphi^T \in \mathbb{R}^{C_{\text{int}}\times 1}$ is also computed using channel-wise 1×1×1 convolutions. $F_t$ represents the feature maps, which are produced by the bottom-up networks (t=1, …, T-1. The total number of stages, T, is set to 4 in our experiments).

We model the feature filter $F_t$ by the relationship between $G_{t+1}$ and $A_t$. The feature tensor $F_t$ and $F_{t+1}$ can be fused only when $G_{t+1}$ is large and $A_t$ is small. Equation 1 is conformed to the feature distribution of different feature layers in deep convolution neural network (Zeiler and Fergus, 2014). Specifically, the deeper the feature layer is, the overall feature (meaning the summary of semantic concept of the image) will have a higher proportion, while detailed features (such as edge, texture) will have a lower proportion. Thus, $(1+G_{t+1})\bullet F_{t+1}$ can capture more overall information of high-level features; on the contrary, $(1+G_{t+1})\bullet A_t \bullet F_t$ is associated more with detailed information of low-level features. Information redundancy can be avoided because the useful information of prediction can be enlarged, while the irrelevant information can be effectively reduced.

3.2.3. Dense top-down feature pyramid module

Decoder module aims to associate more contextual information. Dense connections can further improve the information flow and its propagation between layers(Bilinski and Prisacariu, 2018b; Huang et al., 2017; Lin et al., 2019). Fig. 1 illustrates the connection methods between layers, which referenced the Dense Convolutional Network (DenseNet). Consequently, the feature fusion strategies can be formulated as follows,

$$X_t = \begin{cases} X_T & t = T \\ concat(F_t, X_t) & t < T \end{cases} \tag{4}$$

$$X_t = \begin{cases} X_T & t = T \\ H(\sum_{i=t+1}^{T} X_i) & t < T \end{cases} \tag{5}$$

In the expression above, $\left\{X_t \in \mathbb{R}^{H_t\times W_t\times C_t}\right\}_{t=1}^{T}$ (t = 1, …, T; T = 4) associates the features extracted from both bottom-up feature pyramid module and feature filter module, where the network deepens with the increase of

t. $H_t$, $W_t$ and $C_t$ are the height, width and number of channels of the $t^{th}$ feature maps. The nonlinear transformation $H$ is computed using the combination of $3\times3$ convolutions, ReLU activation function and batchnorm. Specifically, $X_t$ concatenate all higher-level features, $X_i(i>t)$, and the output of feature filter, $F_t$.

3.2.4. Model optimization

Model optimization is driven by a loss function, which was minimized via backpropagation(Rumelhart et al., 1988). In the forward process, the batch normalization and rectified linear units (ReLU) are used in all layers, except for the output layers (softmax units was used in these layers). In addition, to avoid overfitting, L2 regularization (weight decay=0.0005) was applied to permit the weight in model to be closer to 0 and the dropout (probability=0.5, 0.1) is applied in the last layer of the decoder module to effectively delete some units. The proposed network (DFPENet) is trained in an end-to-end manner driven by normalized cross-entropy loss, which is defined as

$$L(y,x,\theta)=-\frac{1}{MN}\sum_{i=1}^{M}\sum_{j=1}^{N}\sum_{k=1}^{K}L(y_i^j=k)\log p_k(x_i^j) \tag{6}$$

Here, $\theta$ represents the parameters of DFPENet; The term M and N are the mini-batch size and the number of pixels in each patch. $L(\bullet)$ is an indicator function, which takes 1 when $y=k$ and 0 otherwise. $x_i^j$ is the $j^{th}$ pixel in the $i^{th}$ patch and $y_i^j$ is the corresponding label of $x_i^j$. In addition, $f(x_i^j)$ represents the output of forward process following the softmax at pixel $x_i^j$, the term of $p_k(x_i^j)$ denotes the probability of the pixel $x_i^j$ belonging to the k-th category, which can be calculated by:

$$p_k(x_i^j)=\frac{\exp(f_k(x_i^j))}{\sum_{n=1}^{K}\exp(f_n(x_i^j))} \tag{7}$$

In the backward process, to train DFPENet in the end-to-end manner, $L(\theta)$ is minimized w.r.t. the DFPENet parameters $\theta$. Specifically, using the stochastic gradient descent (SGD) with Nesterov momentum algorithm (Sutskever et al., 2013) with momentum = 0.9 , initial learning rate = 0.007, and the learning rate schedule employ Polynomial learning rate ($learningRate = InitialLearningRate*(1-\frac{iteration}{\max_iterations})^{power}$) with power=0.9. In the experiments, the network is implemented in the Python deep-learning library PyTorch, on a high-performance computing cluster, with four Tesla K80 12GB GPUs.

3.2.5. Model experimental evaluation

To verify the effectiveness of feature filter module and Dense Top-down Feature Pyramid Module in our network for remote sensing segmentation, we perform several experiments on the ISPRS Vaihingen dataset, and the results are discussed in detail.

**Datasets.** The Vaihingen dataset (ISPRS, 2016a) consists of 3-band IRRG (Infrared, Red and Green) image data acquired by airborne sensors. Overall, there are 33 images with a spatial resolution of 9 cm. The average size of each image is around $2500 \times 2000$ pixels. All images have corresponding ground truth images. Following the setup in the online test, 16 images are used as a training set, while the remaining 17 images (image IDs: 2,4,6,8,10,12,14,16,20,22,24,27,29,31,33,35,38) are used to test our model. All datasets are labeled into the following six classes: impervious surface, building, low vegetation, tree, car and clutter. Then, we randomly sample the $600 \times 600$ patches from the original 33 images. The images are processed at the training stage with the following operations: Normalization, Random Horizontal Flip and Gaussian blur. In the end, 24800 images are generated in the training set and 6205 images in the testing set.

**Pre-trained Datasets (PD).** A larger number of data are extracted from the Vaihingen dataset to create a pre-trained dataset, to further improve the performance of the model. 20 images, including 16 images from the training set and 4 images from the validation set, are used. Furthermore, we randomly sample $600 \times 600$ patches from the 20 images. 26800 images are therefore generated in the pre-trained dataset.

**Implementation.** The proposed DFPENet is initialized with two strategies: the bottom-up feature pyramid module is initialized with ResNet101 pre-trained on ImageNet (Russakovsky et al., 2015); the convolutional layers in feature filter module and dense top-down feature pyramid module are initialized with kaiming uniform distribution. Furthermore, the image patches of size $600 \times 600$ are used as inputs, the mini-batch size is set to 4. The maximum iteration is 62 k. The total training time is about 90 h with one NVIDIA Tesla P80 12GB GPU, and the average testing time of one image ($600 \times 600$) is about 2 s. In addition, the training parameters of the DFPENet model on pre-trained datasets is the same as above.

**Evaluation metric.** To assess the performance of the network, three overall benchmark metrics are used. Specifically, F1 score (F1), overall pixel accuracy (OA) and mean intersection over union (mIoU). F1 and OA are defined as:

$$F_1 = (1+\beta^2)\frac{Precision \times Racall}{Precision + Racall} \times 100\%, \quad \beta = 1 \tag{8}$$

$$OA = \frac{TP+TN}{TP+FP+TN+FN} \times 100\% \tag{9}$$

$$Precision = \frac{TP}{TP+FP} \times 100\% \tag{10}$$

$$Recall = \frac{TP}{TP+FN} \times 100\% \tag{11}$$

Where TP, TN, FP, FN, respectively represent true positive, true negative, false positive and false negative. TPs and TNs denote the real positive sample and negative sample pixels which are found by the model. FNs and FPs denote two different errors, the former represents the missing sample pixels, and the latter represents the wrongly extracted pixels, which are not sampled.

mIoU is defined as:

$$mIoU(P_m, P_{gt}) = \frac{|P_m \cap P_{gt}|}{|P_m \cup P_{gt}|} \times 100\% = \frac{TP}{TP + FP + FN} \times 100\% \tag{12}$$

The term $P_m$ is the set of prediction pixels, $P_{gt}$ is the set of ground truth pixels. $\cup$ and $\cap$ respectively represent union and intersection operations.

**Model analysis.** To analyze the effect of each module in remote sensing segmentation, we use the typical deep learning model Deeplabv3+ (ResNet101) as our baseline. As shown in Table 1, the ResNet101 with Dense Feature Pyramid Module (DFPM), Attention Gate Mechanism (ACM) and Gated Convolution Networks (GCN) outperforms the baseline by a considerable margin in each and every category. It is worth noticing that the baseline model cannot distinguish small objects (such as low vegetation and car) well enough due to the lack of low-level features. DFPM and FFM (including ACM and GCN) can effectively obtain the abilities to understand and fuse multi-scale features, which can achieve a better robustness to the small objects. In addition, Pretrained DFPENet provides a good initialization, which can bring extra $3\% \sim 4\%$ improvement compared to un-pretrained DFPENet. Furthermore, visual results presented in Fig.3 show that our DFPENet model performs better in segmentation and recognition of remote sensing images.

Table 1 Detailed performance comparison of each component in our proposed DFPENet. DFPM: dense feature pyramid module; ACM: Attention Gate Mechanism; GCN: Gated Convolution Networks; PD: Pre-trained Datasets

| Method | Imp. Surf. | Build. | Low veg. | Tree | Car | Mean $F_1$ | mIoU | OA |
|---|---|---|---|---|---|---|---|---|
| ResNet101 (baseline:Deeplabv3+) | 88.52 | 92.26 | 77.29 | 85.80 | 76.77 | 84.13 | 73.09 | 86.20 |
| ResNet101+DFPM | 89.51 | 93.09 | 78.76 | 86.50 | 78.20 | 85.21 | 74.70 | 87.22 |
| ResNet101+DFPM+ACM | 89.70 | 93.39 | 78.65 | 86.11 | 78.65 | 85.30 | 74.83 | 87.26 |
| ResNet101+DFPM+ACM+GCN (DFPENet) | 89.79 | 93.52 | 79.32 | 86.67 | 77.99 | 85.46 | 75.10 | 87.61 |
| ResNet101+DFPM+ACM+GCN (DFPENet+PD) | **92.77** | **95.53** | **86.38** | **90.71** | **81.17** | **89.31** | **81.06** | **91.38** |

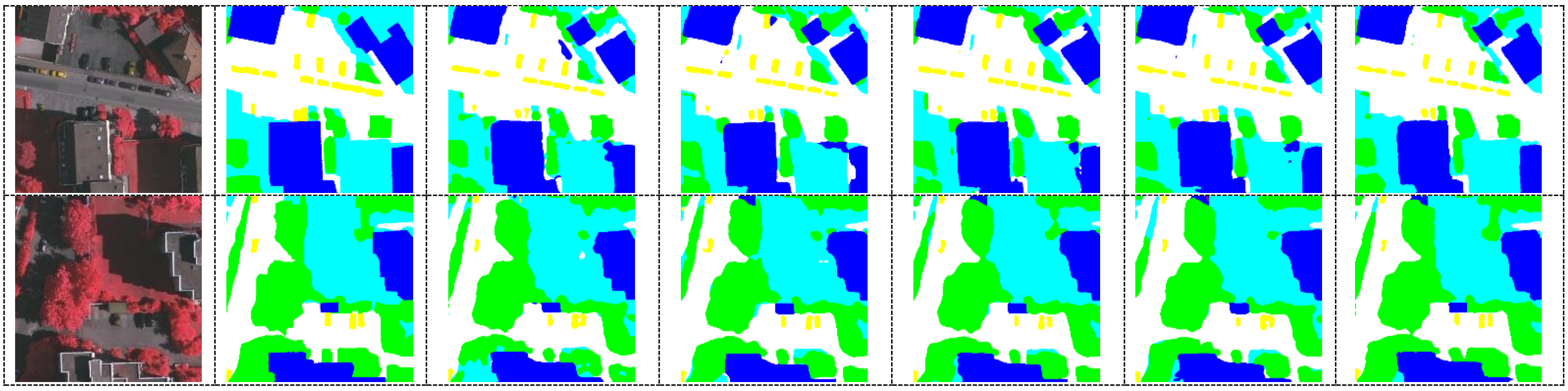

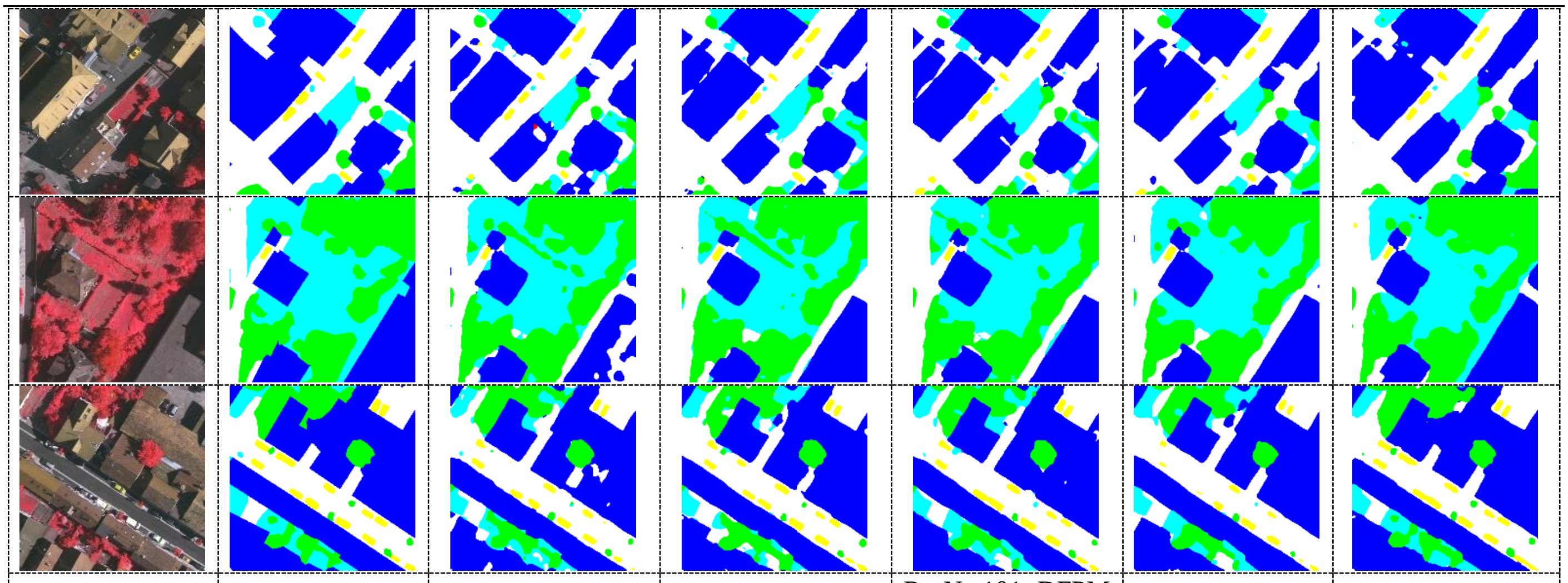

Fig. 3 Examples of segmentation results on the Vaihingen dataset. Legend-white: impervious surfaces, blue: buildings, cyan: low vegetation, green: trees, yellow: cars.

**Benchmark Evaluation.** To further evaluate the effectiveness of the core segmentation model for seismic landslide-DFPENet, comparing with other leading benchmark models in table 2. SVL_6(Gerke, 2014) is one of the baseline methods implemented by the challenge organizer. It utilizes SVL-features(Gould et al., 2009) and a trained Adaboost-based classifier. In addition, a CRF (Conditional Random Field) model is used to fine tune the final prediction. UZ_1 (Volpi and Tuia, 2017) is a CNN (Convolutional Neural Network) model based on encoder-decoder. ADL_3 (Paisitkriangkrai et al., 2016) extracts the features by CNN and from hand-crafted features, which uses CRF as post-processing step. Moreover, DST_2 (Sherrah, 2016) uses a hybrid FCN structure to combine image data with the DSM data, and ONE_7 (Audebert et al., 2016) fuses the output of the two multi-scale SegNets (Badrinarayanan et al., 2017). GSN5 (Wang et al., 2017) utilizes entropy as a gate function to select features. DLR_10 (Marmanis et al., 2016) combines boundary detection with SegNet (Badrinarayanan et al., 2017) and FCN (Long et al., 2014). BKHN11(Minh and Sang, 2018) makes use of FCN (Long et al., 2014). CASIA2 (Liu et al., 2018) uses the self-cascaded network with encoder-decoder. NLPR3(Sun et al., 2018) uses FCN (Long et al., 2014) and a fully connected conditional random fields (F-CRF) is used to further improve the results. HUSTW5(Sun, 2018) utilizes deconvolution network combined with U-Net (Ronneberger et al., 2015). HUSTW5(Sun, 2018) has achieved the state-of-the-art results among all published methods using the Vaihingen dataset. Note that all the metrics are computed using an alternative ground truth in which the boundaries of objects have been eroded by a 3-pixel radius.

As shown in Table 2, the performance of DFPENet model with Pre-trained Datasets (DFPENet+PD) outperforms other advanced models, including HUSTW5. Furthermore, our model performs better than the existing ones in all the given categories, except for the car.

Table 2 Benchmark evaluation results on the ISPRS Vaihingen dataset (class F1 score and overall accuracy), where the values in bold are the best and the values underlined are the second best. Imp surf: impervious surfaces. Low veg: low vegetation. Build: building

| Method | imp. surf. | build. | low veg. | Tree | Car | Overall Acc. |
|---|---|---|---|---|---|---|
| **SVL_6** | 86.00 | 90.20 | 75.60 | 82.10 | 45.40 | 83.20 |
| **UZ_1** | 89.20 | 92.50 | 81.60 | 86.90 | 57.30 | 87.30 |
| **ADL_3** | 89.50 | 93.20 | 82.30 | 88.20 | 63.30 | 88.00 |
| **DST_2** | 90.50 | 93.70 | 83.40 | 89.20 | 72.60 | 89.10 |
| **ONE_7** | 91.00 | 94.50 | 84.40 | 89.90 | 77.80 | 89.80 |
| **GSN5** | 91.80 | 95.00 | 83.70 | 89.70 | 81.90 | 90.10 |
| **DLR_10** | 92.30 | 95.20 | 84.10 | 90.00 | 79.30 | 90.30 |
| **BKHN11** | 92.90 | 96.00 | 84.60 | 89.90 | **88.60** | 91.00 |
| **CASIA2** | 93.20 | 96.00 | 84.70 | 89.90 | 86.70 | 91.10 |
| **NLPR3** | 93.00 | 95.60 | 85.60 | 90.30 | 84.50 | 91.20 |
| **HUSTW5** | 93.30 | 96.10 | 86.40 | 90.80 | 74.60 | 91.60 |
| **DFPENet+PD** | **94.68** | **96.69** | **88.18** | **92.09** | 88.05 | **93.06** |

3.3. DFPENet-Geology: Recognition scheme for co-seismic landslides

In the last decade, there was an increasing interest among the scientists using satellite images in landslide identification (Lu et al., 2019a). Due to the huge amount of data for remote sensing images, it is difficult to optimize the computation speed while keeping it accurate in seismic landslides identification. Thus, we propose a rapid recognition scheme of regional landslides to improve the accuracy of seismic landslide identification while satisfying the requirement of disaster emergency response. In addition, the extracted results are informationized and editable, which can provide reliable information for rapid disaster assessment and emergency response. The proposed scheme consists of four phases, shown in Fig.4.

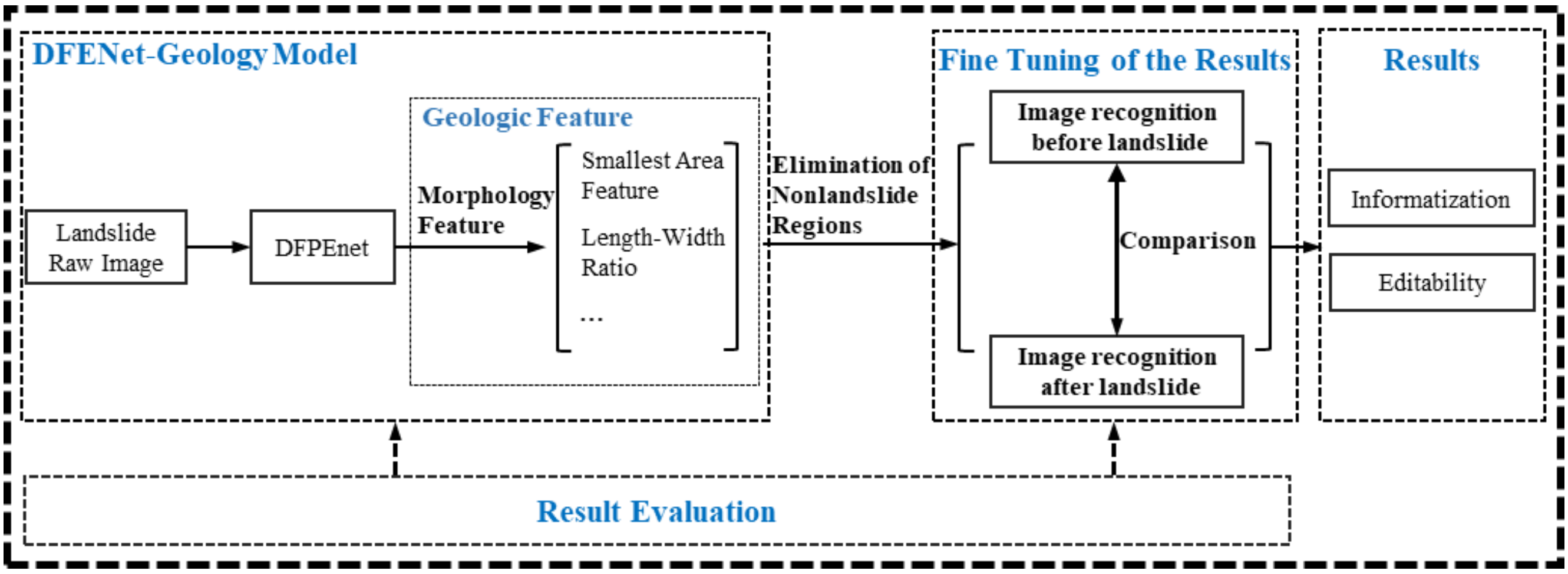

Fig. 4 Process flow of the recognition scheme for co-seismic landslides.

**Phase 1: Training and classification of DFPENet-geology model.**

(1) Model training: Give training set $X$ and corresponding labels $Y$, which can include a small number of landslide samples in the identified area. Additionally, the recommended size of the samples is $512 \times 512$. Furthermore, Section 3.2 can provide a reference for the detailed training parameters.

(2) Classification: It is difficult to classify the whole remote sensing image due to the high spatial resolution, large area, and high complexity of remote sensing images. Therefore, we design a subregion recognition for regional remote sensing images, which is shown in Fig. 5.

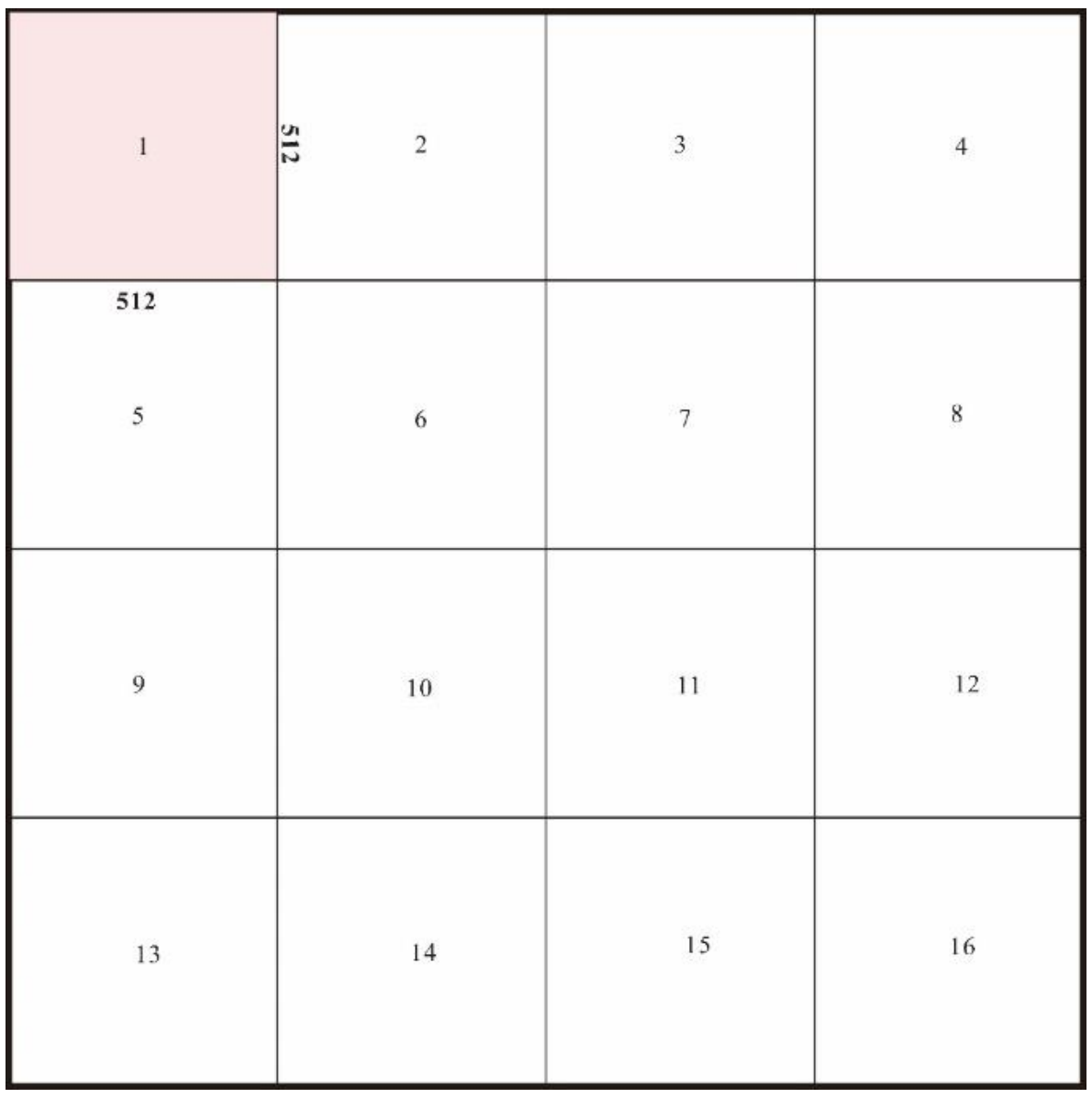

Fig. 5 The diagram of a subregion recognition for regional remote sensing images

(3) Geologic feature fusion.

The input images of DFPENet only contain RGB images of regional landslides, more geological features of landslides are included in the non-RGB data. Thus, it is important to integrate geologic features into the DFPENet. On the contrary, Geologic features of landslides include landslide morphological features, geological features, topographic and geomorphological features, climatic and hydrological features, as well as vegetation features. Generally, we build geologic feature fusion module according to different regions and existing databases. For instance, normalized difference vegetation index (NDVI) (Lu et al., 2019a; Ma et al., 2016; Mondini et al., 2011), topographic feature (slope, aspect, curvature) (Fanos et al., 2018) , morphological feature (Ma et al., 2016), and other geologic features are used for seismic landslide detection and recognition.

We build a geologic feature fusion module with morphological features, due to the limitation of RGB images only. First, the area and length-width ratio of each landslide are obtained by DFPENet segmentation model using Canny edge detection (Canny, 1986). A certain threshold is then established based on the seismic landslide features. Thus, the non-landslide regions segmented by DFPENet model are eliminated. Specifically, the smallest area feature in the remote sensing images is the minimum information about geological hazards that can be extracted directly from the image. Although the ideal landslide list should include all landslides that are possible to detect down to sizes of 1–5 m in length (Harp et al., 2011), these small landslides have little impact on subsequent research. Thus, in subsequent analysis, the minimum landslide area that can be visually interpreted is no less than 4 times its spatial resolution and not more than 25 times its spatial resolution (Ma et al., 2016; Xu, 2015). Length-width ratio is another landslide morphological feature in geologic feature fusion module. The length–width ratios of the landslide are generally quite small in contrast to roads (Barlow et al., 2006; Martha et al., 2010; Orris and Williams, 1984). Therefore, the length-width ratio of the external rectangle of each region could be used to eliminate part of the non-landslide regions (such as roads). The geologic feature fusion module can be formulated as,

$$Y_i = Fusion(X_i + G_i) \tag{13}$$

$$G_i = \begin{cases} Nonlandslide & g_i < A_i \ or \ g_i < T_i \\ Landslide & \text{otherwise} \end{cases} \tag{14}$$

$$T_i = l_i / w_i \tag{15}$$

Where, the term $Y_i$ denotes the $i^{th}$ result of DFPENet-geology model. $X_i$, $G_i$ respectively represent the $i^{th}$ results of DFPENet model and geologic feature fusion module. In addition, $A_i$ represents the minimum landslide area feature, $T_i$ is length-width ratio, which is calculated by the length $l_i$ and width $w_i$ of external rectangle of each region.

**Phase 2: Fine tuning of the model results**

Fine tuning the DFPENet-geology model results by the temporal resolution of the seismic remote sensing images. First, the trained DFPENet-geology is used to sub-regionally recognize the remote sensing images pre- and post-earthquake respectively. Next, two recognition results are subtracted to remove misidentification areas. Finally, the temporal dimension of the regional seismic remote sensing image is added to the model to achieve a high-precision recognition process of the seismic landslide.

**Phase 3: Informatization and vectorization of the results**

In order to further analyze the spatial relationship of regional landslides, the segmentation results of landslides are vectorized and making it editable. The canny edge detection is first used to detect the contour of the segmentation results. The area, perimeter, length-width ratio and centroid of the shape of landslides are then calculated. Furthermore, the identified landslides in the remote sensing images are numbered, and the corresponding geographic coordinates of centroid of landslides are added. These characteristics of landslides are defined as the corresponding field properties, which are summarized in the attached list.

**Phase 4: Evaluation of segmentation model for seismic landslides**

To assess the performance of the segmentation model for seismic landslides, two evaluation schemes can be utilized. One is to evaluate the confusion metric using three benchmark metrics, which is calculated by the validation set of DFPENet-geology model. The correctness (precision) is the fraction of predicted landslide pixels which are labeled as landslide; completeness (recall) is the fraction of all the labeled landslide pixels that are correctly predicted; additionally, mean intersection over union (mIoU) is used to evaluate the DFPENet-geology model. These metrics are defined in section 3.2.5 The other suggests that if the area to be identified has ground truth given in advance (such as the earthquake-triggered Jiuzhaigou landslides in section 4.1), it is necessary to analyze the consistency between automatically identified landslides and actual landslides, considering the pixel number of landslide images and numbers of landslides. In fact, precision, recall and mean intersection over union (mIoU) defined in section 3.2.5 are used to evaluate the consistency after fine tuning the DFPENet-geology model.

### 3.4 Transfer learning for seismic landslide recognition and segmentation

For a learnable and life-long model to perform different seismic landslide recognition tasks, it should be able to reutilize the information acquired in previous seismic landslide segmentation tasks, and transfer it to the new learning tasks of seismic landslide recognition which has only a few samples. Particularly, only a few samples could be selected from each seismic landslide area, due to the time limitation of post-disaster assessment and the insufficiency of landslide data. In this classification settings, training a DNN as well as traditional machine learning approaches may lead to overfitting (Dwivedi and Roig, 2019). One of the solutions to tackle these problems is transfer learning. Transfer learning describes a model that is trained on one data set and then applied to another data set (Oquab et al., 2014).

Given the training set $X_1$ of previous trained model, and the small samples $X_2$ of seismic landslide area to be identified. Transfer learning includes three strategies: (1) Fine-tuning $X_2$: Network is pre-trained on the $X_1$ and later fine-tuned on the $X_2$; (2) Fine-tuning features: Some convolutional layers are fixed or shared in the network and the model trained on the $X_2$ is fine-tuned; (3) Fine-tuning $X_1 + X_2$: Network is firstly pre-trained on the $X_1$, then both $X_1$ and $X_2$ are fine-tuned. However, numerous experiments (Bendale and Boult, 2016; Simonyan and Zisserman, 2014; Sutskever et al., n.d.; Tajbakhsh et al., 2016) have proven that fine-tuning $X_2$ might be one of the most practical and efficient methods because the initialization plays a significant role in transfer learning models, which is adopted in our transfer learning models for seismic landslide recognition and segmentation.

We elaborate the implementation details of transfer learning model in section 4.2, taking the earthquake-triggered Hokkaido landslides in Japan as an example, along with suggestions on the selection of training samples.

## 4. Experimental evaluation

### 4.1. The earthquake-triggered Jiuzhaigou landslides

A small landslide database was established for the experiment on the recognition of earthquake-triggered Jiuzhaigou landslides, and the experimental results was presented in this section. The section described the research regions and implementation details of the segmentation scheme, the scheme results, and its consistency analysis with the artificial interpreted ground truth.

#### 4.1.1. Research regions and data

On August 8, 2017, a magnitude Ms 7.0 earthquake occurred in Jiuzhaigou County, Sichuan Province, China. Its epicenter was located at 33.20° N and 103.82° E with a focal depth of 20 km. By the evening of August 13, there were 25 deaths with 525 injuries, 76,671 damaged houses and a total estimated economic loss of more than 140 million RMB (The People’s Government of Sichuan Province, 2017). Moreover, the earthquake had a tremendous impact on the Jiuzhaigou Scenic Area. Numerous earthquake-triggered landslides had caused at least 29 road obstructions and damages in the scenic area. The total length of the damaged road was about 4 km. For reconstruction of the scenic area, it was critical to investigate the spatial state of these landslides. The research regions were chosen in the Jiuzhaigou Scenic Area, covering an area of 53.6 $km^2$, which is shown in the Fig. 6. In addition, the pre- and post-earthquake remote sensing images were from 1.5 m-resolution satellite image (Google Earth, December 2016) and 1.5 m-resolution UAV photographs, respectively.

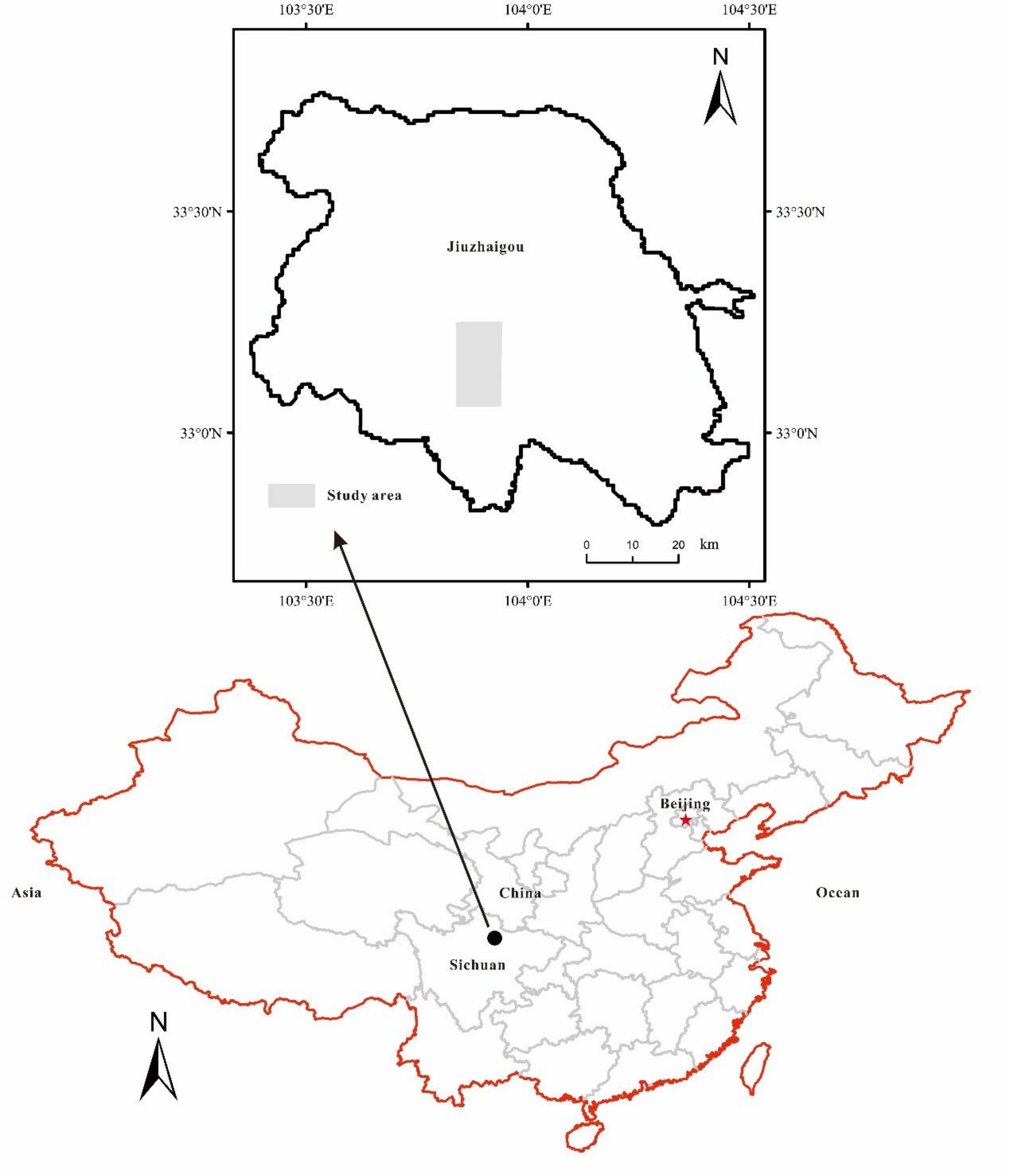

Fig. 6 The geographic location of the research region.

Table 3 The detailed information of the landslide database.

| Category | Data source | Data description | Amount | Total |
|---|---|---|---|---|
| Training set | Google Earth | Landslides triggered by the 2008 Wenchuan earthquake and 2017 Jiuzhaigou earthquake (less than 5m resolution) | 576 | 950 |
| | UAV | 2015 Lishui landslide, 2017 Maoxian landslide, 2018 Xishan Village landslide, Jiuzhaigou seismic landslides around Xiongmaohai and other typical landslides (less than 2m resolution) | 384 | |
| Validation set | Google Earth | Landslides triggered by the 2008 Wenchuan earthquake (less than 5m resolution) | 50 | 70 |
| | UAV | Lishui landslide, Jiuzhaigou seismic landslides (2m resolution) | 20 | |

A landslide database was established by a set of optical images from Google Earth, UAV photographs, and other data sources. The landslide database was then divided into training and validation set. It is worth noticing that in order to enhance the feature of the Jiuzhaigou landslides, the area centered on Xiongmaohai was also added to the training set (Fig. 6). In addition, we processed the images with random horizontal and vertical flip operations to enlarge the database. Then, the landslide database was expanded to 1020 optical images, each image being $512\times512$ pixels. The ground truth of all these images are available. Detailed information of the database were summarized in the Table 3.

4.1.2 Experimental results

**Implementation details.** The proposed DFPENet was initialized by ResNet101 which was pre-trained on ImageNet (Russakovsky et al., 2015). The image patches of size $512\times512$ were used as inputs, the mini-batch size was set to 4. In addition, the maximum iteration was 30k, and the total training time is about 27.2 h with one NVIDIA Tesla P80 12GB GPU, and the average testing time of a post-earthquake image ($5176\times10349$) was about 120 s by subregion recognition (230 times of scans in total), while the testing time of a pre-earthquake image ($4658\times10282$) was about 110 s (scanned 209 times in total). It was noticeable that the histogram matching should be performed when there was a big difference of histogram between pre- and post-earthquake images.

**Results of the research region.** The post-earthquake image was identified by the trained DFPENet model. The reference landslide superimposed on the post-event image was presented in Fig. 7(b), and 712 landslides were identified in the post-earthquake image. Because the study area was dominated by small and medium landslides and the impact of the roads could be eliminated by processing the pre- and post-earthquake images, the feature of the smallest area was then considered as the geological limitation in the geologic feature fusion module (the minimum area was 37.5 $m^2$), thus 573 landslides were obtained (Fig. 7(c)). Next, the temporal resolution was added. The pre-earthquake image was identified by subregion recognition, and then its identification results was subtracted from the counterpart of the post-earthquake image. Lastly, 522 landslides were obtained (Fig. 7(d)) in the research region. The result was then vectorized and editable, and the corresponding attributes and numbers were added to obtain the results in Fig. 8. The recognized landslides can be accessed in appendix A.

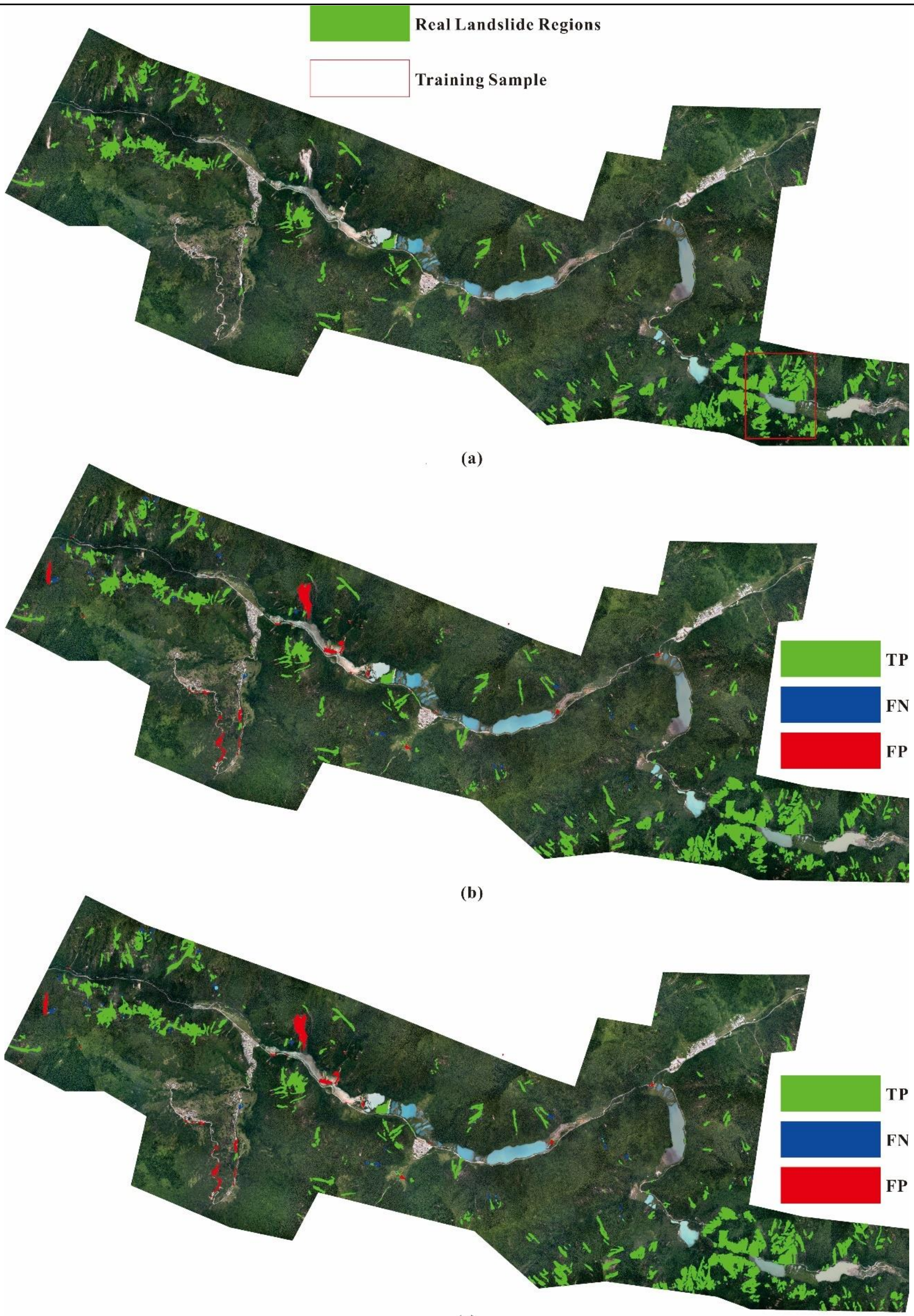

(a)

(b)

(c)

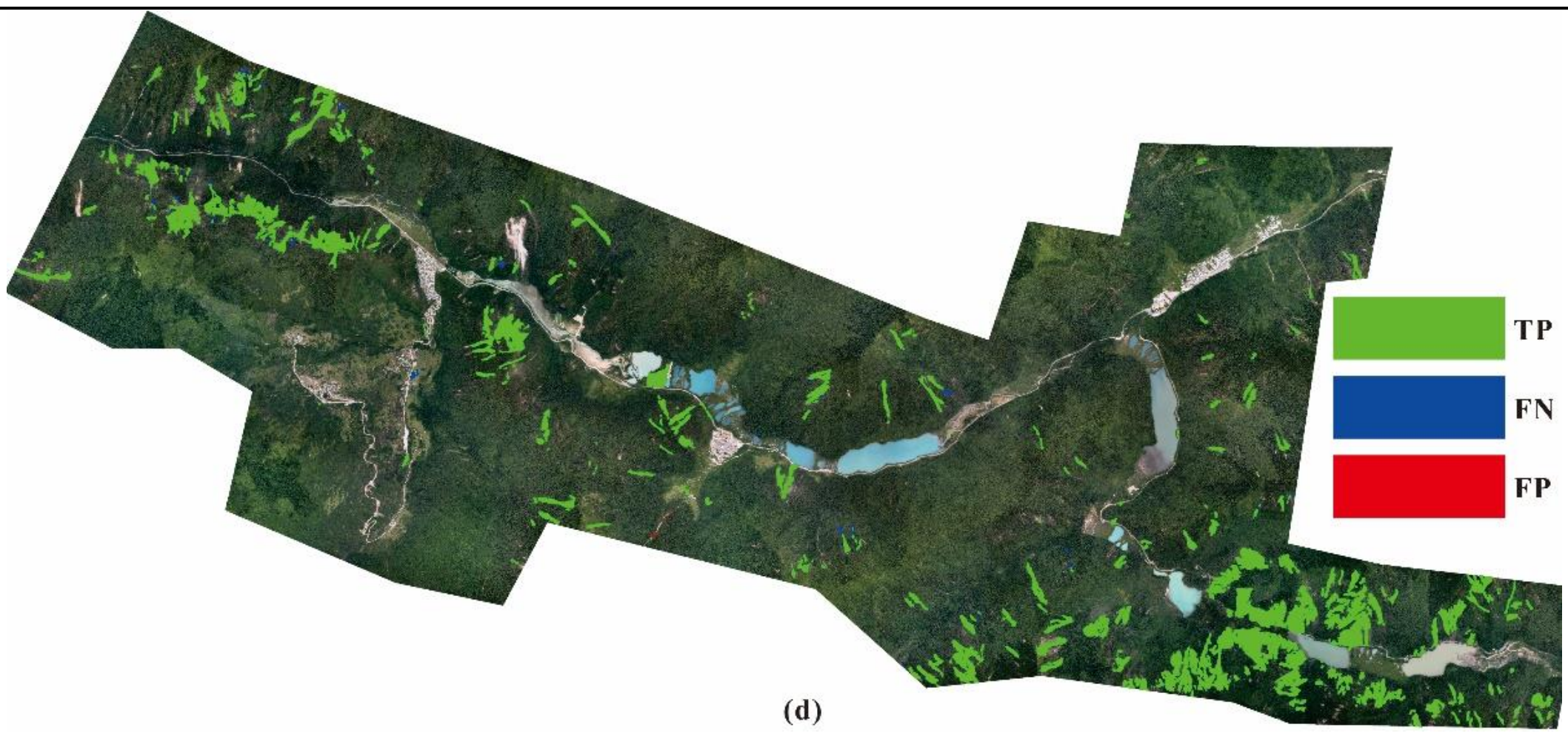

Fig. 7 (a) the ground truth of the research region; (b) the results of DFPENet model; (c) the results of elimination of non-landslide regions by geologic feature；(d) the results of eliminating non-landslide regions by comparison between pre- and post-event. In the colored figures, True Positive (TP) is marked in green, False Positive (FP) in red, and False Negative (FN) in blue.

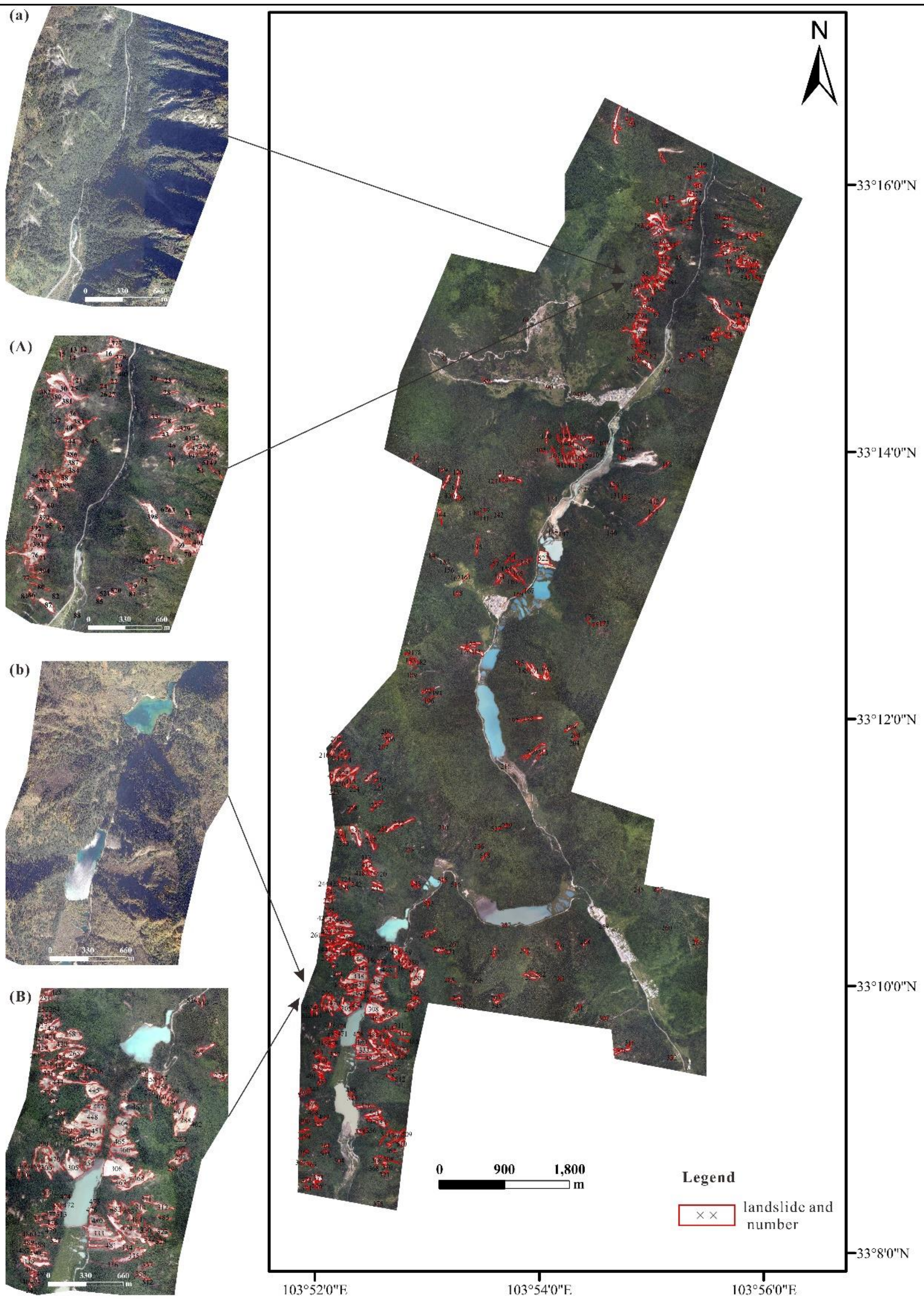

Fig. 8 Informatization and vectorization of the final results

**Evaluation.** Three benchmark metrics were used to evaluate the confusion metric which was calculated by the validation set of DFPENet model, which was shown in the second line of the Table 4. On the other hand, in order to further evaluate the performance of recognition model in Jiuzhaigou seismic landslides, it was necessary to analyze the consistency between automatically identified landslides and actual landslides by calculating the pixel number of landslide images and number of landslides. Notably, the ground truth (Fig. 7(a)) of research region was mainly obtained by artificial visual interpretation, in addition, a few Jiuzhaigou seismic landslide investigations could provide references for determining landslide boundaries (Fan et al., 2018).

Furthermore, the experimental results of the proposed method are concluded in Table 4, which was based on the pixel number of landslide image. For the numbers of landslide areas, the results of accuracy assessment were shown in Table 5.

According to the consistency analysis results, the final comprehensive performance of our identification model for regional landslides could reach to 98.67%, without considering the landslide boundary error on the pixel number of landslide image. Currently, the best accuracy of landslide identification by object-oriented multi-scale segmentation could reach about 85% (Ma et al., 2016), the accuracy of geological disaster recognition based on deep auto-encoder with wavelet coefficient was up to 97.40%, while generalization ability of the model is not strong enough (Liu and Wu, 2016). The landslide mapping precision (correctness) of the Change Detection-based Markov Random Feld (CDMRF) approach were greater than 0.75 in single-sensor data tests and over 0.70 in double-sensor data test (Lu et al., 2019b). In addition, the accuracy of landslides detection based on DCNN was up to 80.01% (Zhong Chen et al., 2018). Therefore, our experimental results showed that the proposed model has set a new state-of-the-art performance on the regional landslide identification. Furthermore, according to the Table 4, the accuracy could be increased by 0.08% on the pixel number of landslide images by using the smallest landslide area feature, while the accuracy can be improved by 6.39% by comparing the recognition results between pre- and post-earthquake. This result showed that the comprehensive performance of the regional landslide recognition model could be further improved by considering the temporal resolution.

According to the consistency analysis on the number of landslides in Table 5, the accuracy of DFPENet model and DFPENet-geology model was improved from 67.44% to 82.06%. It was observed that the DFPENet model could easily misunderstand many small exposed bedrocks as landslides in this test area. Furthermore, according to the number of TP areas, the results of DFPENet model and the fine-tuned varied from 526 to 519. It was observed that the DFPENet model could easily segment some of the large landslides into multiple parts, resulting in errors in the number of landslides. In general, the model performance reached to 89.95% in the number of landslides.

Overall, based on the comparison between Table 4 and Table 5, although our method has obtained high recognition accuracy on the pixel number of landslide images, certain defects in the recognizing the landslide boundary accurately still existed. These defects could be further improved by the geologic feature fusion module.

Table 4 The experimental results based on the validation set and the pixel number of landslide image.

| Category | model | Number of TPs | Number of FPs | Number of FNs | Precision | Recall | mIoU (Accuracy) |
|---|---|---|---|---|---|---|---|
| Validation set | DFPENet | - | - | - | 91.09% | 91.46% | **83.95%** |
| | DFPENet | 915698 | 55853 | 20753 | 94.25% | 97.78% | 92.28% |
| Consistency analysis | DFPENet -geology | 915698 | 55040 | 20753 | 94.33% | 97.78% | 92.36% |
| | Fine tuning results | 924252 | 245 | 12199 | 99.97% | 98.70% | **98.67%** |

Table 5 The experimental results based on the numbers of landslide areas.

| Category | model | Number of TP Areas | Number of FP Areas | Number of FN Areas | Precision | Recall | mIoU (Accuracy) |
|---|---|---|---|---|---|---|---|
| Consistency analysis | DFPENet | 526 | 186 | 68 | 73.88% | 88.55 % | 67.44% |
| | DFPENet -geology | 526 | 47 | 68 | 91.80% | 88.55 % | 82.06% |
| | Fine tuning results | 519 | 3 | 55 | 99.43% | 90.42 % | 89.95% |

4.2 The earthquake-triggered Hokkaido landslides in Japan

This section presented the experimental results of transfer learning for Hokkaido landslide recognition. The following section elaborated on the implementation of transfer learning, the optimal number of small training sets for landslide recognition models, and the results.

4.2.1 Research regions and data

On September 6, at 03:08 a.m. JST, a magnitude $M_w$ 6.6 earthquake stuck the Iburi-Tobu area of Hokkakido, Japan. Its epicenter located at 42.72° North, and 142.0° East, and the depth of the hypocenter was approximately 37.0 km (Japan Meteorological Agency (JMA), 2018). Several studies have stated that the earthquake occurred due to the effect of powerful typhoon Jebi and the region's complex tectonic setting (Kobayashi et al., 2019; Wang et al., 2019; Yamagishi and Yamazaki, 2018). Importantly, the seismic event then triggered landslides, affecting an area of 700 $km^2$. The earthquake reportedly caused more than 44 deaths and over 660 injuries, more than 80% of the casualties were caused by landslides. Therefore, it is of significant importance for disaster relief to recognize co-seismic landslides with high efficiency and high precision. We selected the research regions centered on the epicenter of the earthquake, whole area covering 159.3 $km^2$ (Fig. 9). The pre- and post- earthquake remote sensing images were downloaded from the Planet respectively on August 3 and September 11 of 2018 with the resolution of 3 m.

A small dataset was selected from the post-earthquake remote sensing image, which was then divided into a training set and a validation set. Specifically, tile A ( $910\times2206$ ) was set as the training set, while tile B ( $1221\times1496$ ) was used as the validation set (Fig. 9). Two strategies for selecting data were listed as follows: The selected landslide features were representative in the entire area; the pixel ratio between the landslide and the background should be as balanced as possible. Following the previous settings, we randomly sampled $512\times512$ patches from the small dataset. This led to 100 images in the training set and 30 images in the validation set. In addition, the images were processed at the training stage by Normalization, Random Horizontal Flip and Gaussian blur.

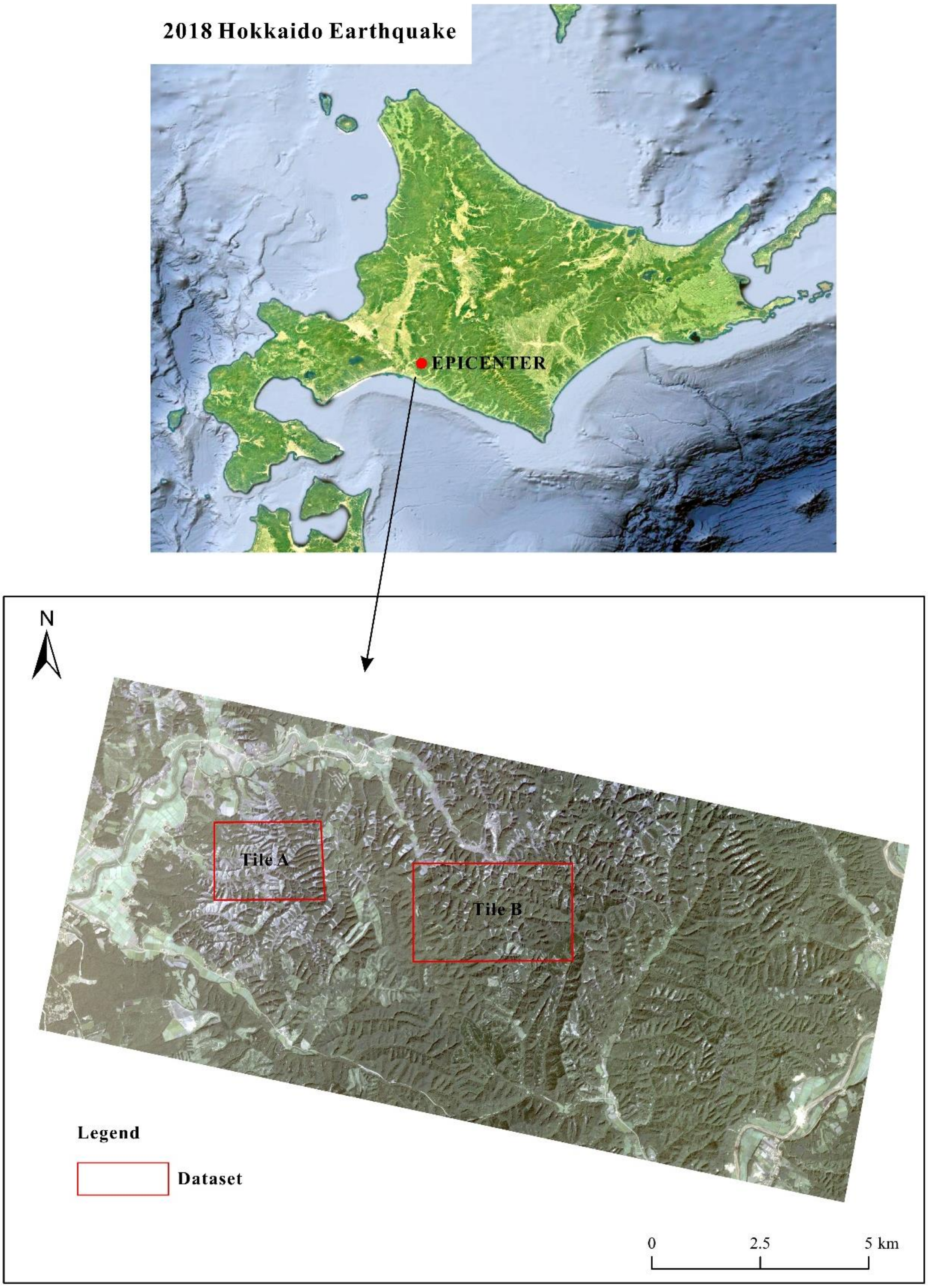

Fig. 9 The RGB image of post-earthquake acquired on 11 September 2018, illustrating the geographic location of the research region.

4.2.2 Experimental results

**Implementation details.** The Jiuzhaigou landslides recognition model described in section 4.1 was fine tuned. The maximum iteration was 1k, and the total training time is about 80 m with one NVIDIA Tesla P80 12GB GPU, and the average testing time of post-earthquake image ($8983 \times 4516$) was about 90 s by subregion recognition (161 times of scanned in total), while the testing time of pre-earthquake image ($8956 \times 4535$) was also about 90 s.

**Size of training set.** With the interest of how training set size affect the performance of transfer learning method, we carried out the following experiments. We respectively fine-tuned the model with 5%, 10%, 20%, 40%, 60%, 80% and 100% of the training set (containing 100 images, i.e. N=100), with the mini-batches set to 2,2,4,4,4,4,4 respectively, while evaluating the performances in the validation set (N=30). According to the three benchmark metrics: mIoU, precision and recall (Fig. 10), the model performance raised following the increase of data amount in the training set. Specifically, to ensure the model accuracy to exceed 60%, the training set size must be no less than 20%, preferably more than 40%. When the sample size was too small (less than 20%), the experimental results showed that the mIoU/precision was low, and the recall was high, which indicated that the model mistakenly identified some of the landslides as the background. Therefore, it was necessary to consider the pixel weight of the landslide sample, to increase the model performance in this situation. What we used for the fine-tuned model in Hokkaido landslide recognition was the full training set (N=100). The specific results of the model in the validation set was shown in Table 6.

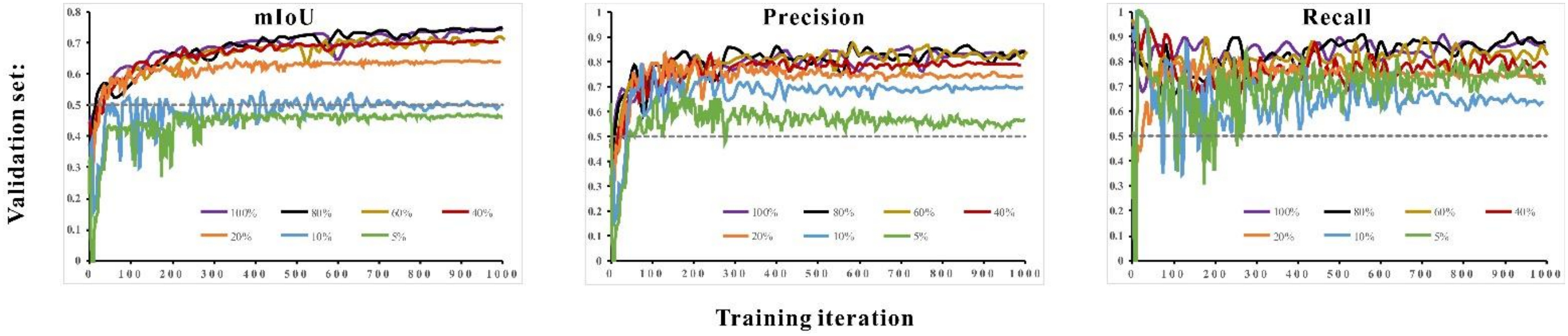

Fig. 10 The results of three benchmark metrics, when the model is fine-tuned with 5%, 10%, 20%, 40%, 60%, 80% and 100% of the training set (N = 100)

Table 6 The experimental results of the fine-tuned model in Hokkaido landslide recognition.

| Model | Precision | Recall | mIoU |
|---|---|---|---|
| DFPENet | 84.28% | 90.20% | 77.21% |

**Results of study area.** The pre-earthquake image (Fig. 10(a)) and post-earthquake image (Fig. 10(c)) were identified respectively by the fine-tuned DFPENet model. The reference landslide (blue) superimposed on the post-event image was presented in Fig. 10(d), and 1,987 landslides were identified in the post-earthquake image. The feature of smallest area was considered as a geological limiting factor in the geologic feature fusion module (the minimum area was 75 $m^2$) because roads had less impact on the recognition results, which was shown in the black pixels in the Fig. 10(e). Furthermore, the temporal resolution was added into consideration. According to the pre-earthquake identification results (Fig. (b)), partial non-landslide regions were eliminated, which was shown in the red pixels in the Fig. 10(e). In total, 1,362 landslides were obtained (Fig. 10(f)) in the research region. The result was then vectorized and editable, and the corresponding attributes and numbers were added to obtain the result in Fig. 11. The recognized landslides can be accessed in appendix B.

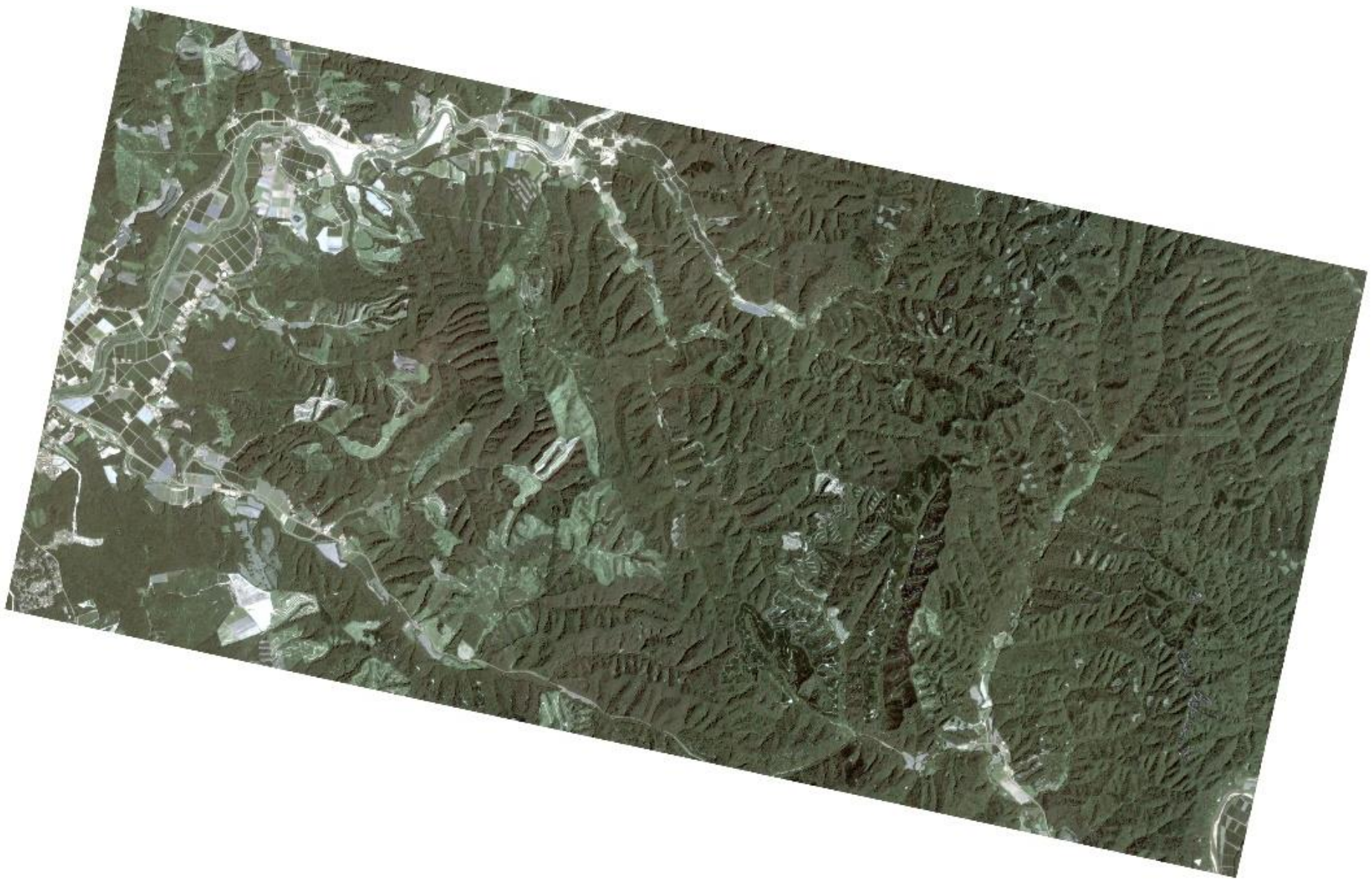

**(a)**

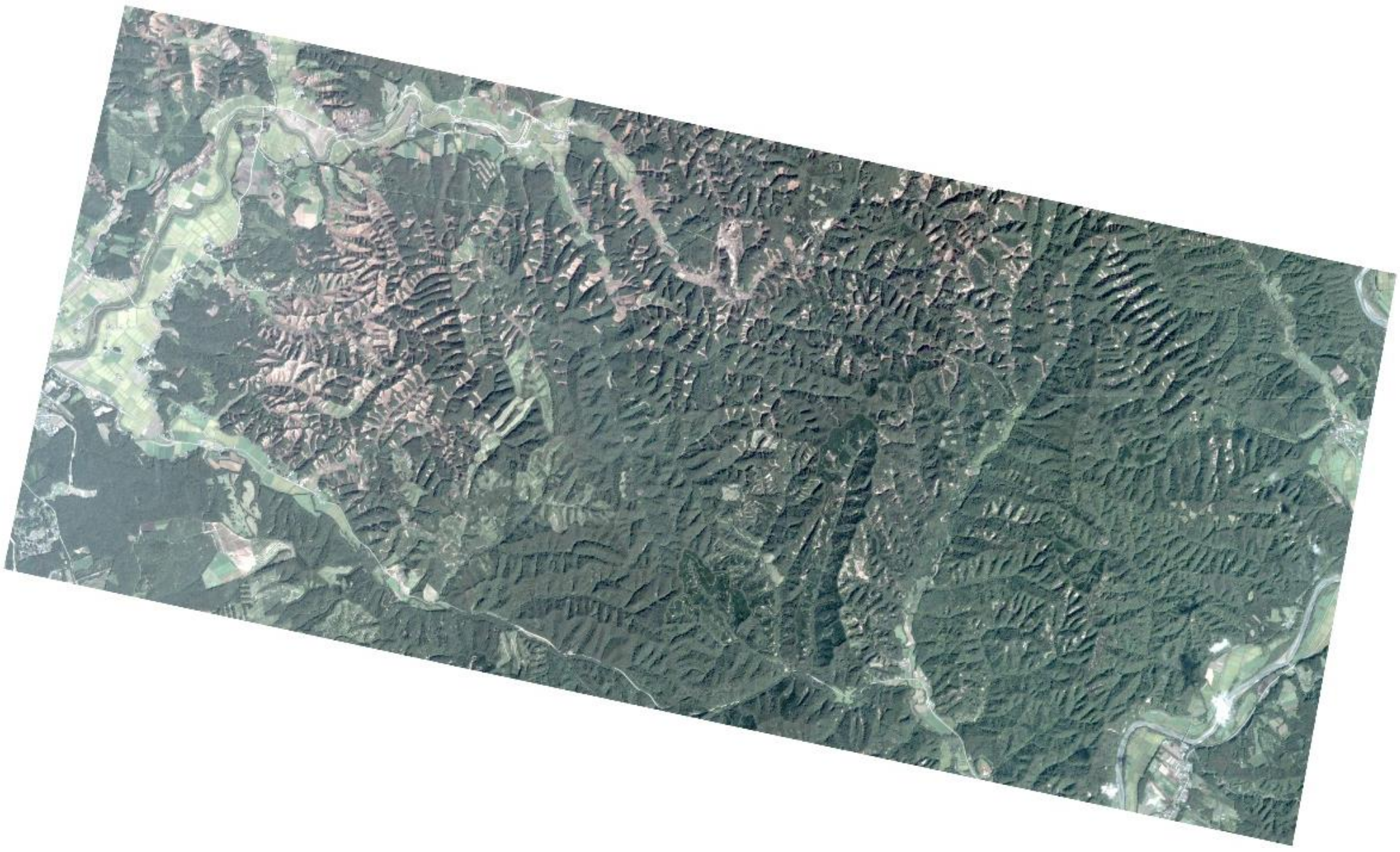

**(b)**

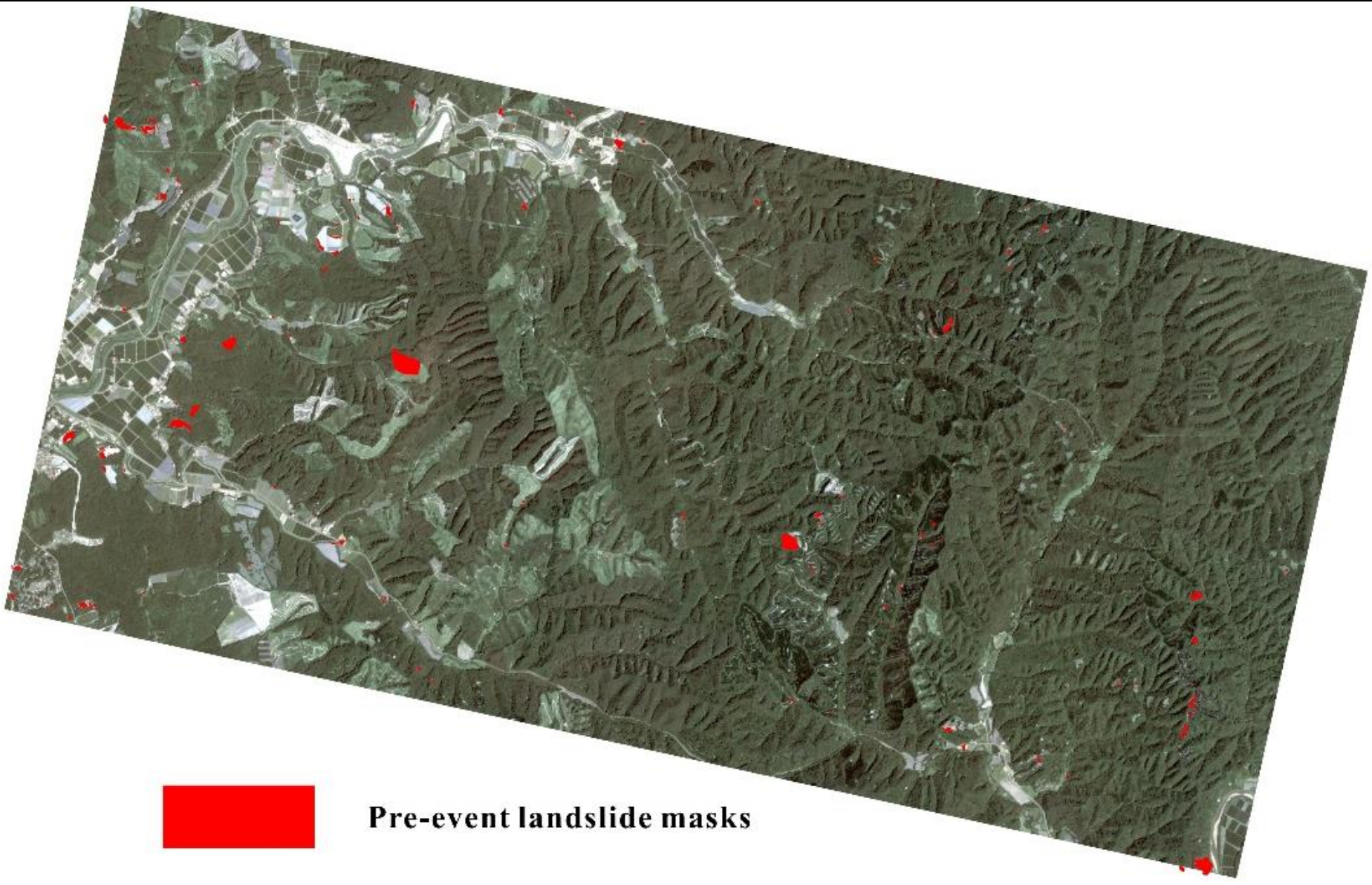

(c)

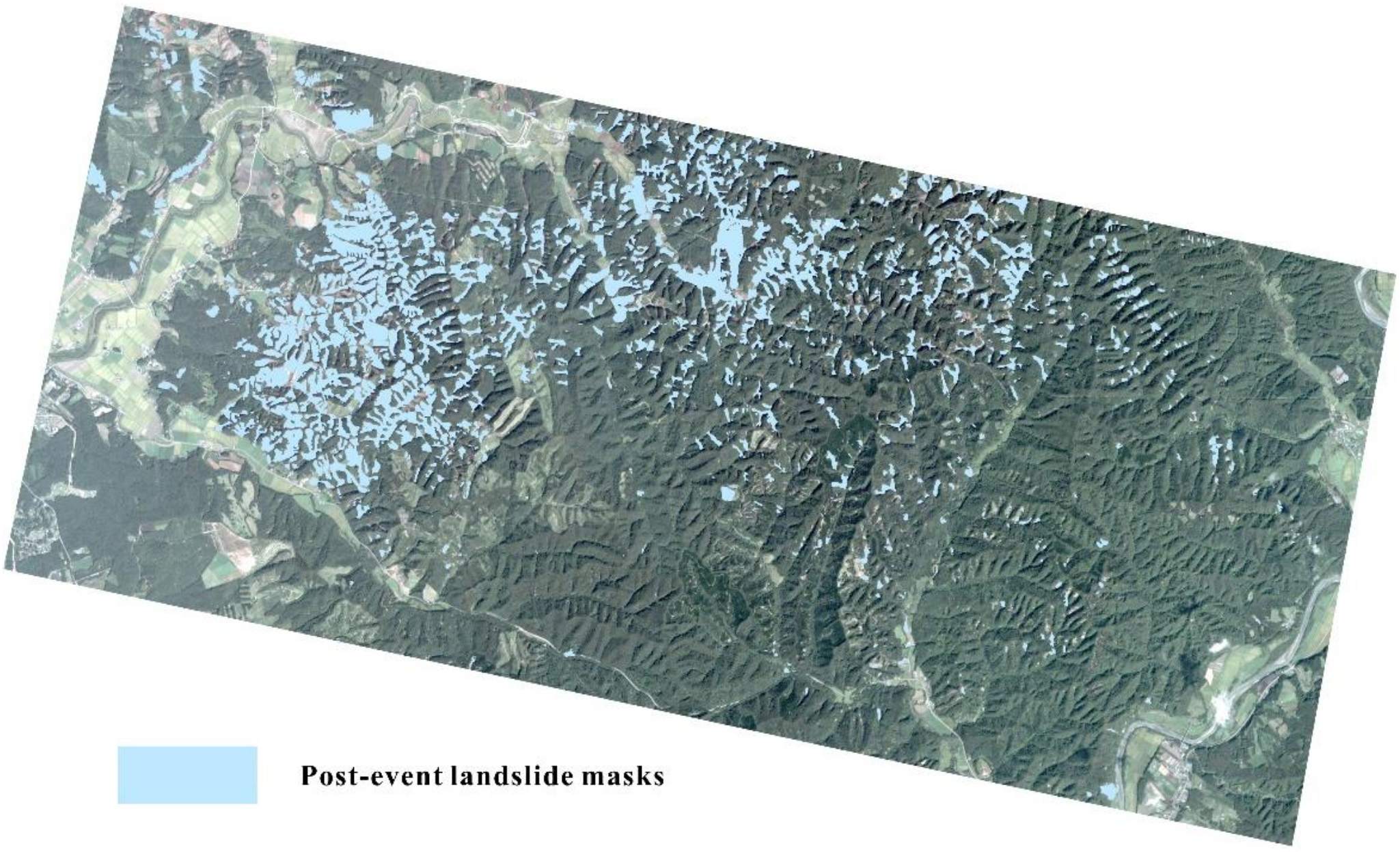

(d)

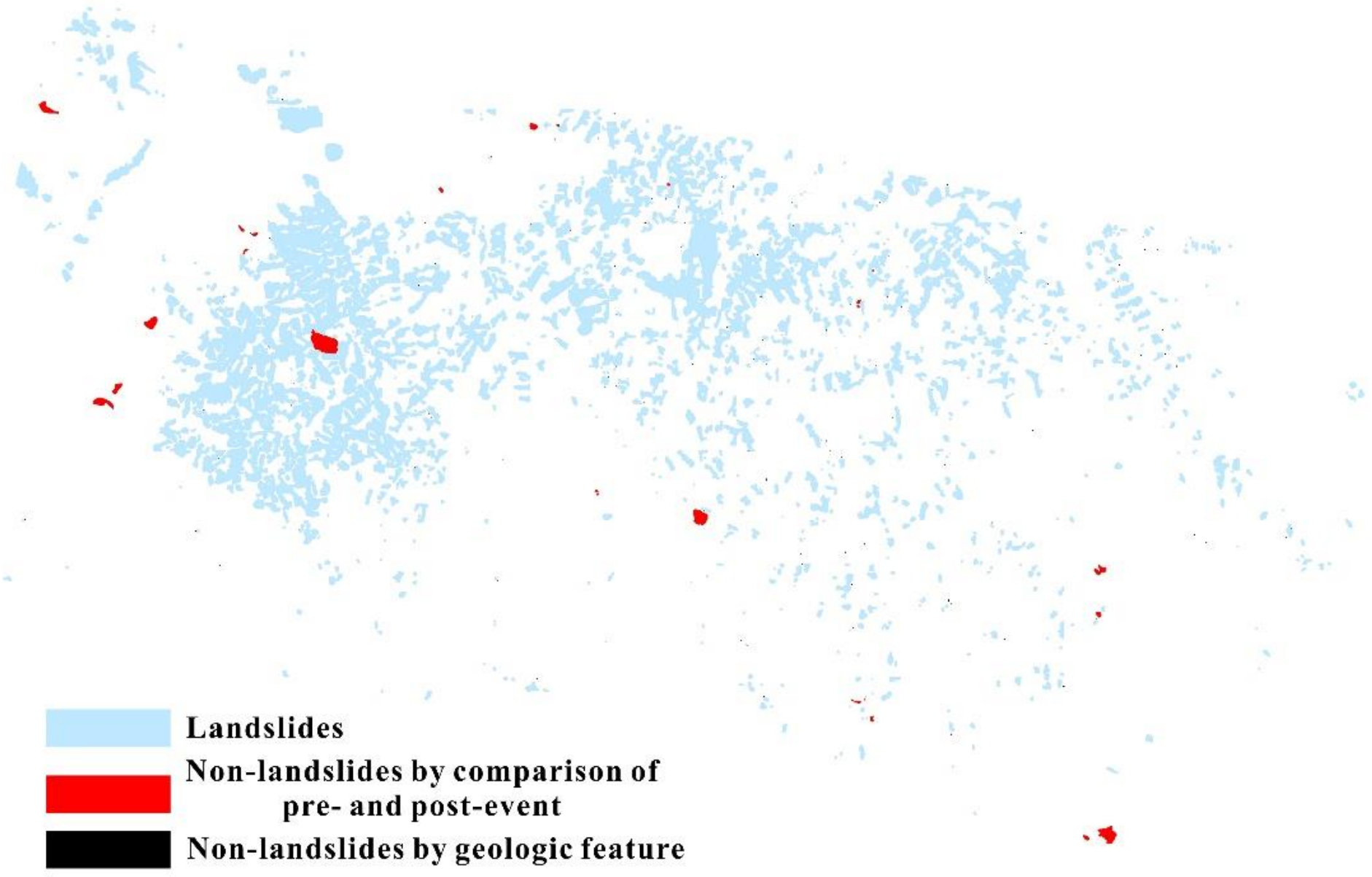

(e)

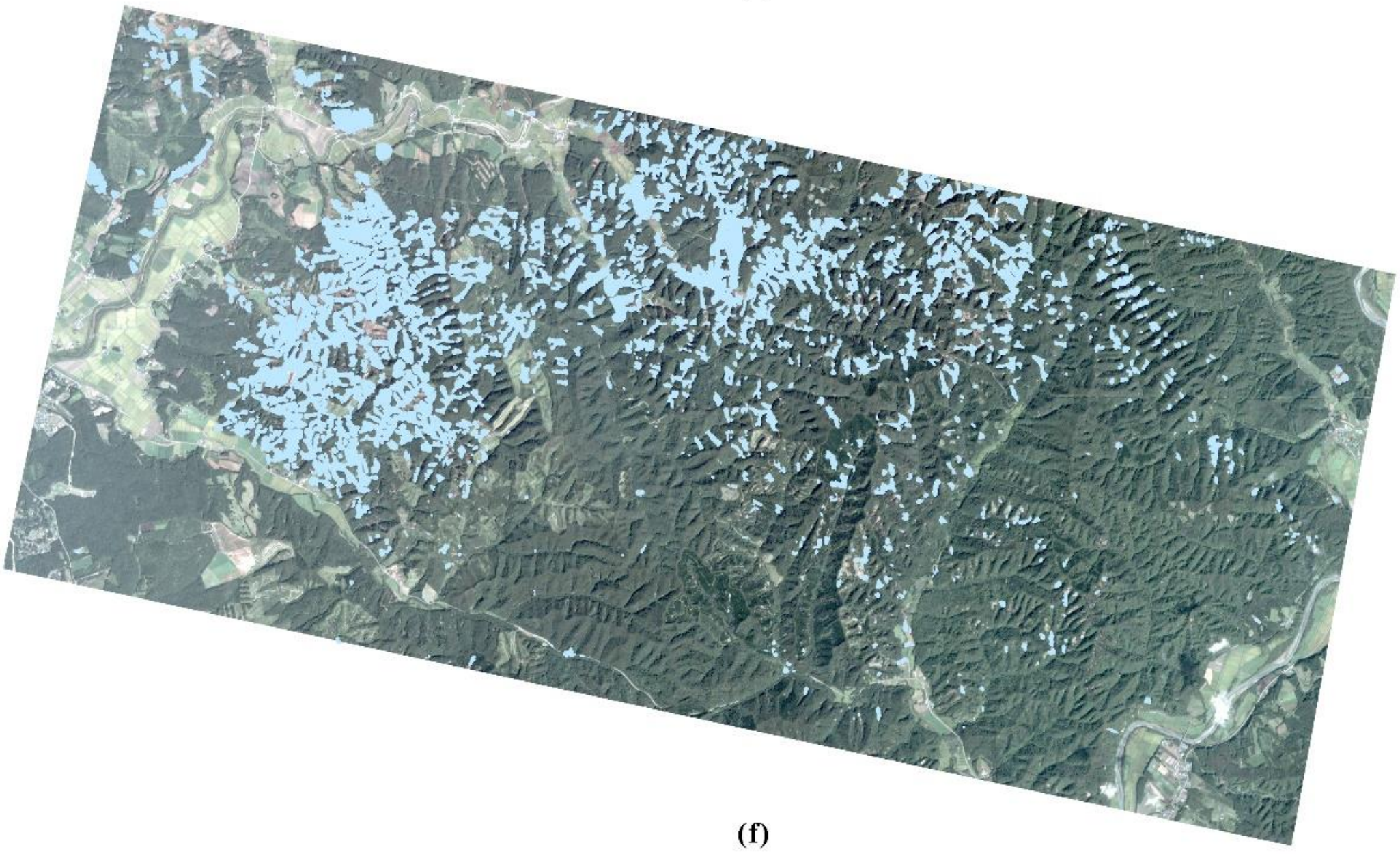

(f)

Fig. 10 (a) the remote sensing image of pre-earthquake; (b) the remote sensing image of post-earthquake; (c) the reference landslide (red) superimposed on the pre-event image; (d) the reference landslide (blue) superimposed on the post-event image; (e) the results of elimination of non-landslide regions by geologic feature and comparison between pre- and post-event; (f) the final identification results superimposed on the post-event image

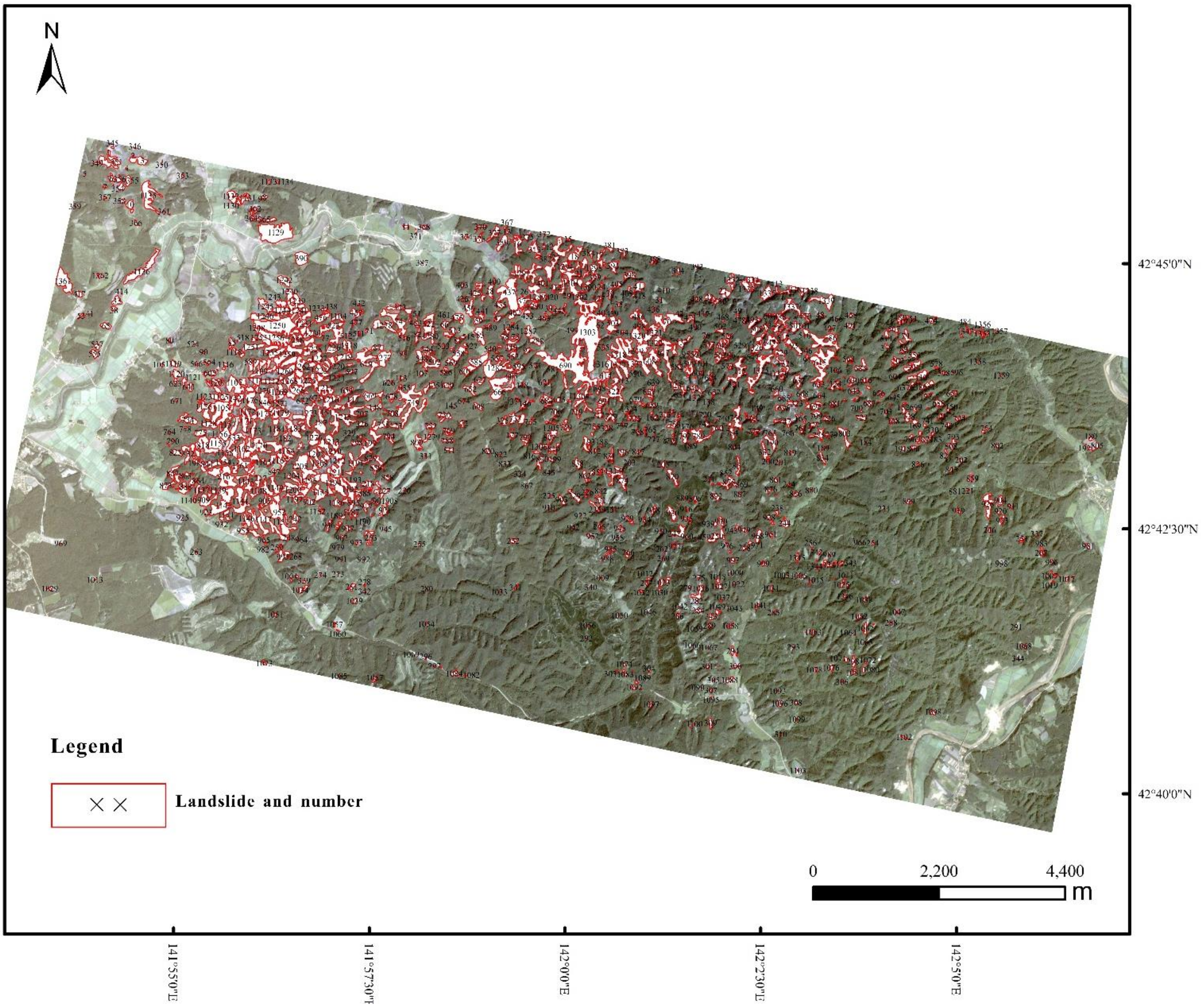

Fig. 11 Informatization and vectorization of the final results

## 5. Conclusions

This study develops a robust model for deep convolutional neural network research in remote sensing. It may establish an accurate, rapid, comprehensive, and cross-scene scheme for seismic landslide recognition, by integrating image features extracted from the DFPENet model, geologic features, temporal resolution, landslide spatial analysis, and transfer learning, while only RGB images of seismic remote sensing are used in the scheme. First, Dense Feature Pyramid with Encoder-decoder Network (DFPENet) is proposed in this paper to understand and fuse the multi-scale features of objects in remote sensing images. In this model, a new feature filter module is proposed, which can also be adopted in other deep learning frameworks. In addition, by using ISPRS Vaihingen datasets, the local evaluation results show that our segmentation model outperforms other state-of-the-art deep learning models. Secondly, a comprehensive and widely-used recognition scheme for seismic landslides, based on DFPENet-geology and transfer learning, is designed. To corroborate its feasibility and applicability, the proposed scheme is applied to the earthquake-triggered

Jiuzhaigou landslides in China and the earthquake-triggered Hokkaido landslides in Japan, with high-resolution remote sensing images before and after the earthquakes. The experiment results show that the proposed scheme also presents a new state-of-the-art performance in regional landslide identification while not considering the landslide boundary error, and the scheme performs well in different seismic landslide tasks. Future research of our team will focus on further integration of deep learning methods and seismic hazards, bringing more prior knowledge of seismic landslides into deep neural networks, to realize an end-to-end recognition of earthquake landslides.

## Model and code availability

Algorithms, DFPENet models (including Un-pretrained and Pretrained DFPENet models), as well as recognition models of co-seismic landslides (including the model trained with landslide database and the model transferred to the earthquake-triggered Hokkaido Landslides) are publicly available on Github under a GNU General Public License (https://github.com/xupine/DFPENet). The landslide database will be available in due course. In addition, utilizing our schemes to accomplish recognition tasks for other seismic landslides is highly welcomed.

## Acknowledgments

Financial support was provided by the Strategic Priority Research Program of CAS (Grant No. XDA23090303), the National Key Research and Development Program of China (Project No. 2017YFC1501000), the NSFC (Grant No. 41572303), the CAS Youth Innovation Promotion Association.